\documentclass{article} 
\usepackage[table]{xcolor}   
\usepackage{style_file,times}

\usepackage{amsmath,amsfonts,bm}

\def\eqref#1{equation~\ref{#1}}

\def\1{\bm{1}}

\DeclareMathAlphabet{\mathsfit}{\encodingdefault}{\sfdefault}{m}{sl}
\SetMathAlphabet{\mathsfit}{bold}{\encodingdefault}{\sfdefault}{bx}{n}

\usepackage{hyperref}
\usepackage{url}

\usepackage[utf8]{inputenc} 
\usepackage[T1]{fontenc}    
\usepackage{hyperref}       
\usepackage{url}            
\usepackage{booktabs}       
\usepackage{amsfonts}       
\usepackage{nicefrac}       
\usepackage{microtype}      

\definecolor{gold}{rgb}{1.0, 0.84, 0.0}
\definecolor{silver}{rgb}{0.75, 0.75, 0.75}
\definecolor{bronze}{rgb}{0.8, 0.5, 0.2}

\usepackage{graphicx}
\usepackage{amsmath}
\usepackage{cleveref}
\usepackage{amsthm}
\usepackage{bbm}
\usepackage{subcaption}
\usepackage{dsfont}
\usepackage{enumitem}

\newtheorem{theorem}{Theorem}[section]

\newtheorem{definition}[theorem]{Definition}

\newtheorem{axiom}[theorem]{Axiom}

\newcommand{\smallerpm}{\text{\scriptsize$\pm$~}}

\usepackage{xspace}

\newcommand{\FU}{\texttt{FairlyUncertain}\xspace}

  \usepackage{nth}
  \usepackage{intcalc}

  \newcommand{\cAAAI}[1]{AAAI\ Conference\ on\ Artificial (AAAI)}

\title{FairlyUncertain: A Comprehensive Benchmark of Uncertainty in Algorithmic Fairness}

\author{Lucas Rosenblatt\thanks{Truly equal contribution, in alphabetical order} \hspace{.1em} \& R. Teal Witter$^{*}$\\
New York University \\
\texttt{\{lucas.rosenblatt,rtealwitter\}@nyu.edu} \\
}

\iclrfinalcopy 
\begin{document}

\maketitle

\begin{abstract}
Fair predictive algorithms hinge on both equality and trust, yet inherent uncertainty in real-world data challenges our ability to make consistent, fair, and calibrated decisions. While fairly managing predictive \textit{error} has been extensively explored, some recent work has begun to address the challenge of fairly accounting for irreducible prediction \textit{uncertainty}. However, a clear taxonomy and well-specified objectives for integrating uncertainty into fairness remains undefined. We address this gap by introducing \FU, an axiomatic benchmark for evaluating uncertainty estimates in fairness. Our benchmark posits that fair predictive uncertainty estimates should be \textit{consistent} across learning pipelines and \textit{calibrated} to observed randomness. Through extensive experiments on ten popular fairness datasets, our evaluation reveals: (1) A theoretically justified and simple method for estimating uncertainty in binary settings is more consistent and calibrated than prior work; (2) Abstaining from binary predictions, even with improved uncertainty estimates, reduces error but does not alleviate outcome imbalances between demographic groups; (3) Incorporating consistent and calibrated uncertainty estimates in regression tasks improves fairness without any explicit fairness interventions. Additionally, our benchmark package is designed to be extensible and open-source, to grow with the field. By providing a standardized framework for assessing the interplay between uncertainty and fairness, \FU paves the way for more equitable and trustworthy machine learning practices.
\end{abstract}

\section{Introduction}
Fairness in machine learning enhances transparency and trust in algorithmic predictions, and is both a legal and moral imperative given the direct impact predictive models can have on peoples' lives. Although extensive research has addressed reducing predictive error disparities among demographic groups -- by tackling limited data, model biases, or structural inequities -- achieving fairness also necessitates accurately assessing the uncertainty associated with each prediction; however, the precise interplay between algorithmic fairness and uncertainty estimates remains an open question \citep{bhatt2021uncertainty, hendrickx2024machine}.

Assessing the interplay between fairness and uncertainty is challenging due to a lack of principled objectives for integrating the two notions. Much work has contended with how to define and measure fairness in predictive models \citep{lee2018detecting,caton2020fairness,mitchell2021algorithmic, barocas2023fairness,corbett2023measure}.
Additionally, predictions made on real-world data have inherent \textit{uncertainty}, given factors like noisy individual behavior, measurement errors, and environmental influences. In statistics and machine learning, a careful typology helps organize these sources of randomness. Uncertainty or variance that is specific to each observation is said to display \textit{heteroscedasticity} (as opposed to \textit{homoscedasticity}, which is not a function of the observation itself). Heteroscedastic uncertainty can either be \textit{epistemic} or \textit{aleatoric} in nature \citep{hullermeier2021aleatoric}. Epistemic uncertainty is uncertainty that can be reduced with more data \citep{chen2018my}; while aleatoric uncertainty is uncertainty that arises from the inherent data distribution, and cannot be reduced with larger samples. We will focus on models that provide a single uncertainty estimate for each observation-specific prediction, but will carefully reason about the \textit{aleatoric} component of that uncertainty estimate, with the goal that this quantity is \textit{consistent with} and \textit{calibrated to} the data. 

\begin{figure}[h!]
    \centering
    \vspace{-1em}
    \includegraphics[scale=.5]{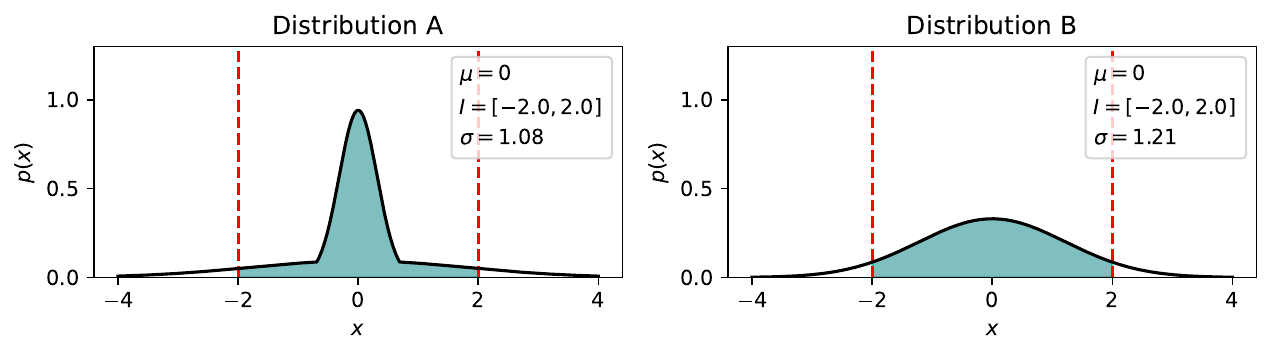}
    \vspace{-1.3em}
    \caption{Two distributions over observable outcomes. For example, Distribution A can represent the test scores of a student in a stable home whereas Distribution B can represent the test scores of a student in an unstable home. While both have the same mean and $80\%$ confidence interval, the distributions are substantially different as captured by the standard deviation. }
    \label{fig:distributions}
\end{figure}

As a concrete example, consider the task of predicting student performance on a standardized exam: one student reliably earns the same score on the test while another student, perhaps because of their home situation, earns a significantly different score depending on unobservable factors in their daily life. In Figure~\ref{fig:distributions}, we hypothesize about how differing unseen environmental factors might affect outcome variance. We can produce more accurate \textit{epistemic} uncertainty estimates by incorporating more data and improving our predictive model. However, if we cannot have the same student take the test multiple times, we cannot precisely estimate \textit{aleatoric} uncertainty because we lack repeated observations of a student's performance measuring individual fluctuations under identical conditions. 

\begin{table}[t!]
    \centering
    \caption{The following \textit{uncertainty typology} allows us to speak precisely about sources of uncertainty we can estimate directly from data, and sources of uncertainty we can only hope to approximate.}
    \label{tab:uncertainty_typology}
    \resizebox{\linewidth}{!}{
    \begin{tabular}{p{5cm} p{6cm} p{6cm}}
        \toprule
        \midrule
        \textbf{Uncertainty Type} & \textbf{Description} & \textbf{Example} \\
        \midrule
        \textbf{(A) Unmeasurable individual-level uncertainty} & Uncertainty inherent to individual outcomes that cannot be measured. & A student may experience a unique, random moment of distraction during a test. \\
        \midrule
        \textbf{(B) Within-individual variability} & Uncertainty captured through repeated measurements of an individual. & Scores for one student on tests throughout the year across different testing conditions. \\
        \midrule
        \textbf{(C) Across-individual variability} & Uncertainty arising from differences between individuals (covariate attributable). & Two students perform differently due to factors such as access to study resources, etc. \\
        \midrule
        \textbf{(D) Sampling uncertainty} & Uncertainty stemming from repeated sampling i.e. process of data collection. & Differing test scores between two random samples of students from sample variability. \\
        \midrule
        \textbf{(E) Modeling uncertainty} & Uncertainty introduced by the modeling process itself (hyperparameters, etc.). & Different models give slightly different predictions for the same set of students. \\
        \bottomrule
    \end{tabular}}
    \vspace{-1.5em}
\end{table}

Its worth noting that estimating forms of heteroscedastic uncertainty -- uncertainty at the individual level -- in the context of algorithmic fairness has received substantial recent interest \citep{liu2022conformalized, ali2021accounting, han2022umix, tahir2023fairness, wang2024aleatoric}. 
Some approaches use ensembles of models to estimate uncertainty while others train models to learn the uncertainty directly.
In both cases, the models are trained through a machine learning pipeline on a particular architecture with specific hyperparameters.
Unfortunately, the estimates can vary substantially with respect to a model's depth, activation function, and other settings \citep{lakshminarayanan2017simple,guo2017calibration,malinin2018predictive}. As methods for uncertainty estimation in prior work differ on the actual estimation objective (to varying degrees), we find it useful to provide a precise typology over sources of uncertainty \citep{hullermeier2021aleatoric} in Table~\ref{tab:uncertainty_typology}. This typology helps us to define precise targets for each uncertainty estimation evaluation strategy in our benchmark.

As an example of how this typology helps describe reasonable uncertainty estimate objectives, consider the scenario given in Figure~\ref{fig:distributions}, which exemplifies both \textbf{(A)} and \textbf{(B)} uncertainty in Table~\ref{tab:uncertainty_typology}. Producing a \textit{fair} uncertainty estimate ultimately lies in understanding \textbf{(A)} and \textbf{(B)} uncertainties \citep{black2021selective,cooper2024arbitrariness}, but empirical methods over a fixed data sample can only directly estimate \textbf{(C)}, \textbf{(D)}, and \textbf{(E)}. A goal of our work is to use estimates of \textbf{(C)}-\textbf{(E)} to assess the \textit{consistency} and \textit{calibration} of approximations for the unmeasurable uncertainties \textbf{(A)} and \textbf{(B)}. In the next section, we specify \textit{consistency} and \textit{calibration} as \textit{axiomatic principles} for successful and fair uncertainty estimates; these principles guide our construction of the \FU benchmark.

\section{Preliminaries}
In Section~\ref{sec:axioms}, we state guiding axioms for \textit{consistent} and \textit{calibrated} uncertainty estimates, as evaluated by the \FU benchmark. First, however, we must define a \textit{learning pipeline} (given in Definition~\ref{def:learning_pipeline}, related to definitions from \citep{black2021selective, long2024individual}), and what makes two learning pipelines \textit{similar} (given in Definition~\ref{def:sim_learning_pipelines}).

\begin{definition}[Learning Pipeline]\label{def:learning_pipeline}
Let $\mathcal{T}$ be a training set of $n$ observations with covariates $\mathbf{x}_1, \ldots, \mathbf{x}_n \in \mathcal{X}$, protected attributes $a_1, \ldots, a_n \in \mathcal{A}$, and outcomes $y_1, \ldots, y_n \in \mathcal{Y}$.
Consider a set of $m$ hyperparameters $\lambda_1, \ldots, \lambda_m$.
A \textit{learning pipeline} $\mathcal{P}$ is a randomized training procedure that takes the training set and hyperparameters to learn a predictive function $f: \mathcal{X} \times \mathcal{A} \to \mathcal{Y} \times \mathbb{R}$. 
The predictive function produces an estimate of the outcome $\hat{y}$ and uncertainty $\sigma \geq 0$ for given covariates $\mathbf{x}$ and protected attributes $a$.
\end{definition}

\begin{definition}[Similar Learning Pipelines]\label{def:sim_learning_pipelines}
For each hyperparameter $\lambda_j$, let $\tau_k$ represent a threshold that takes on a \textit{reasonable} value with respect to hyperparameter $j$.
Two learning pipelines $\mathcal{P}$ and $\mathcal{P}'$ (Definition~\ref{def:learning_pipeline}) are considered \textit{similar} if they differ only in hyperparameter settings i.e., there is some $j \in [m]$ for which $\lambda_k = \lambda_k’$ except when $j = k$ and $|\lambda_j - \lambda_j’| \leq \tau_j$.
\end{definition}

\subsection{Axioms} \label{sec:axioms}

The \FU benchmark focuses on two properties that uncertainty estimates produced by \textit{similar learning pipelines} (Definitions~\ref{def:learning_pipeline} and~\ref{def:sim_learning_pipelines}) should satisfy. Let $f(P)$ be a predictive function from a learning pipeline $P$ such that $f(P): (\mathbf{x}, a) \mapsto (\mu, \sigma)$ i.e. $f(P)$ maps inputs $\mathbf{x}$ and protected attributes $a$ to a point prediction $\mu$ and an estimated standard deviation of the inherent \textbf{(A)} and \textbf{(B)} types of uncertainty $\sigma$ for the predicted value of $\mu$. Note if $f(P)$ is a classification model, $\mu$ is a probability, and if $f(P)$ is a regression model, $\mu$ is a real value. 

Now, we are ready to state Axiom~\ref{ax:consistency}, which formalizes the idea that uncertainty estimates should be a function of the data rather than of an arbitrary learning pipeline. In other words, \textit{similar learning pipelines applied to the same dataset should produce similar uncertainty estimates.}

\begin{axiom}[Consistency]\label{ax:consistency}
For $f$ and $f'$, which are produced by similar learning pipelines, $\sigma$ and $\sigma'$ should be close. This means that if two learning pipelines are similar, the uncertainty estimates they produce should not vary much.
\end{axiom}

This idea is related to the motivation behind selective ensemble and self-consistency ensemble methods \citep{black2021selective,cooper2024arbitrariness}. However, Axiom~\ref{ax:consistency} is a property of uncertainty estimates rather than outcome predictions.

Consistency by itself is not sufficient; a model that always returns 0 is consistent but not meaningful. This leads to our second property, which posits that uncertainty estimates should also be calibrated. In other words, \textit{uncertainty estimates should always explain observed heteroscedastic variance.}

\begin{axiom}[Calibration]\label{ax:calibration}
The predicted uncertainty estimates should satisfy $\textrm{Var}(y~|~\mathbf{x},a) =\sigma^2$, where $\sigma$ is the estimated standard deviation of the outcome $y$ given covariates $\mathbf{x}$ and protected attributes $a$. This means the predicted uncertainty $\sigma$ should match the actual variability observed in the data, capturing the heteroscedastic nature of the uncertainty. When uncertainty estimates are biased or incorrect on average, they fail to be calibrated.
\end{axiom}

Once uncertainty estimates are \emph{consistent} and \emph{calibrated}\footnote{We use \emph{consistent} and \emph{calibrated} to denote the formal/axiomatic definitions throughout, as opposed to common usage. Additionally, we note that Axioms~\ref{ax:consistency} and~\ref{ax:calibration} contain specific assumptions that dictate strategies for evaluation; different formalizations would lead to different strategies. We acknowledge this limitation, but argue that any alternative strategies for evaluating uncertainty should intuitively align with each desiderata.}, the next challenge is figuring out how to integrate them into fair algorithms effectively.
Integrating uncertainty in social settings is commonly done through \textit{abstention}; for a sample where the level of uncertainty is too high, defer making a prediction, thereby avoiding unreliable results \citep{black2021selective,cooper2024arbitrariness}. 
Another method (commonly adopted in regression settings) is to return the uncertainty estimates alongside the predictions, providing a clearer picture of the confidence in the results and allowing practitioners to make more informed decisions \citep{liu2022conformalized,wang2024aleatoric}. In either setting, no consensus best approach for incorporating uncertainty estimates into fairness interventions has emerged; this motivates our benchmark, \FU.

\section{\FU : A Benchmark for Uncertainty in Fairness}\label{sec:fu}

\textbf{Contributions}~~~To the best of our knowledge, \FU is a first of its kind fairness \textit{and} uncertainty benchmark. Prior work has either provided singular uncertainty evaluations~\citep{cooper2024arbitrariness} or extensive fairness benchmarking \textit{without} methods to incorporate  uncertainty~\citep{bird2020fairlearn, aif360-oct-2018}. \FU fills the gap, introducing a variety of tests and experiments, and is designed to be extensible and grow with the literature.
\FU supports five binary datasets and five regression datasets off-the-shelf that are standard to the fair machine learning context; Table~\ref{table:datasets} (deferred to the Appendix) provides prediction task summaries for each. All of our code for \FU is ready for release as a Python package; the benchmark is modular, so that uncertainty methods and datasets can be easily added.
The package includes functions to generate all of our experiments with just a few lines of code. 

\textbf{Road Map}~~~In Section \ref{sec:consistency_calibration}, we begin by evaluating \textit{consistency} and \textit{calibration} on both classification and regression tasks. We find that baseline ensemble methods are the most \textit{consistent}. Additionally, we find that in classification tasks, complex ensemble-based methods from prior work on uncertainty are neither \textit{consistent} nor \textit{calibrated}. In contrast, methods that directly learn uncertainty parameters via negative log-likelihood are both \textit{consistent} and \textit{calibrated}. Next, we incorporate uncertainty estimates into making fair decisions. In Section~\ref{sec:abstaining}, we show that while abstaining from predictions can improve accuracy, this approach does not reduce the imbalance in outcomes between demographic groups. In Section~\ref{sec:fairness_regression}, our results lead us to advocate for outputting both predictions and uncertainty estimates in regression tasks. To facilitate this, we introduce \textit{Uncertainty-Aware Statistical Parity} (Definition~\ref{def:uasp}), a natural generalization of statistical parity in the regression setting that includes uncertainty estimates. Remarkably, we find that \textit{consistent} and \textit{calibrated} uncertainty methods can reduce \textit{Uncertainty-Aware Statistical Parity} without \textit{any} explicit fairness interventions.

\textbf{Experimental Details}~~~We use a cluster of 24-core Intel Cascade Lake Platinum 8268 chips to run the experiments.
We use default model hyperparameters except when the experiment explicitly varies them i.e., for assessing \textit{consistency}.
Our benchmark considers a privileged class and an unprivileged class for each dataset. The privileged class can be at the intersection of multiple protected features; for example, “white” (race) and “male” (gender) could become a privileged class subset of “white males” for evaluation. All fairness metrics in the submitted paper are then binary and are the absolute difference between the disadvantaged and advantaged group. The benchmark defaults to using the XGBoost model for all predictive tasks; XGBoost is a fast, state-of-the-art model that generally outperforms neural models in relatively low data, low dimensional tabular regimes, like most datasets in our benchmark \citep{chen2016xgboost, grinsztajn2022tree, mcelfresh2024neural}. 

\section{Consistency and Calibration}\label{sec:consistency_calibration}

In this section, we evaluate methods for capturing heteroscedastic uncertainty according to Axioms~\ref{ax:consistency} and~\ref{ax:calibration}. \FU provides an evaluation strategy for \emph{consistency} that investigates the sensitivity of heteroscedastic uncertainty estimates to hyperparameters in the learning pipeline (Figure~\ref{fig:binary_consistency}). For learning pipelines that vary only in one hyperparameter, we compute uncertainty estimates and then the \textit{standard deviation} of these estimates (Table~\ref{tab:binary}).

Measuring \textit{calibration} is much harder since we cannot observe more samples from the same distribution (type \textbf{(A)} uncertainty) and would need generated data for type \textbf{(B)}. Thus, we devise strategies which measure uncertainty types \textbf{(C)}, \textbf{(D)} and \textbf{(E)} to \textit{approximately} test \textit{calibration} in a \textbf{qualitative} and \textbf{quantitative} manner. \textbf{Qualitatively}, \FU measures \textit{calibration} over \textit{similar groups,} based on the following intuition: if estimated uncertainties are \emph{calibrated}, observations with higher estimated uncertainties should have higher variation in the test data.
Concretely, we create groups with similar uncertainty estimates; then, within each group, we compute the empirical standard deviation of the residual difference between true and predicted outcomes (Figure~\ref{fig:binary_calibration}).

Now, imagine we have access to the true probability $p_i$ that individual $i$ will belong to the positive class in the binary setting. Then, following the binomial distribution, the true uncertainty as measured by variance would be $\sigma^2 = p_i (1-p_i)$. With our \textbf{quantitative} strategy for measuring \textit{calibration}, we contend that if we have a prediction for the true probability  $\tilde{p}_i \approx p_i$ then a natural prediction for the uncertainty as measured by variance is $\tilde{\sigma}^2 = \tilde{p}_i (1-\tilde{p}_i)$. A canonical method to evaluate the goodness of fit given probabilistic assumptions is \textit{Negative Log Likelihood} (NLL) \cite{fisher1925theory}. 

This leads to the quantitative evaluation given in Table~\ref{tab:binary}, where we make this assumption on the \textit{meaning} of uncertainty estimates in binary predictions to offer a clear and simple measure of \emph{calibration} based on the NLL of observing the data according to the estimated parameters of the Binomial distribution, interpreting estimate $\mu$ as Binomial probability $p$ (with uncertainty estimate $\sigma$ being the standard variance $p(1-p)$). Because there are existing methods which produce uncertainty estimates that cannot be interpreted as standard deviations (namely \cite{black2021selective,cooper2024arbitrariness}), when assessing these methods, one should focus on the output of the \textbf{qualitative} assessment (Figure~\ref{fig:binary_calibration}), which provides a more general purpose approach for approximately testing \textit{calibration}. 

\subsection{Models and Classification Tasks}\label{cc_classification}

We subject several methods for heteroscedastic uncertainty estimation on binary classification tasks to our evaluation.
Many of the methods we consider rely on an ensemble of $k$ models, where each model in the ensemble is trained on samples taken randomly with replacement from the training set.
For predictions, these methods simply output the mode prediction of the $k$ models, while for uncertainty estimates, they differ in their calculations.
We'll let $k^{(0)}$ be the number of models predicting 0 and $k^{(1)}$ be the number of models predicting 1.
Our experiments are run with the following methods:
\begin{itemize}[leftmargin=*]
    \item The \textit{Ensemble} computes the standard deviation of the $k$ predictions to estimate uncertainty.
    \item The \textit{Selective Ensemble} method of \cite{black2021selective} estimates uncertainty by computing the $p$-value of observing the number of negative predictions $k^{(0)}$ and the number of positive predictions $k^{(1)}$ were they sampled from a binomial distribution with probability $\frac12$.
    \item The \textit{Self-consistency Ensemble} method of \cite{cooper2024arbitrariness} estimates uncertainty by computing a so-called self-consistency metric $1-2k^{(0)}k^{(1)}/(k(k-1))$. Since the self-consistency metric increases with certainty, we will report self-(in)consistency, i.e., $2k^{(0)}k^{(1)}/(k(k-1))$ for parity with the other approaches.
    \item The \textit{Binomial NLL} method contrasts with the ensemble methods by directly estimating uncertainties under the Binomial assumption e.g. by producing probabilities $p$ that minimize negative log likelihood on the training set, yielding an uncertainty estimate that is the Binomial standard deviation $\sigma = \sqrt{p (1-p)}$. We offer a brief formal justification as to why we expect \textit{Binomial NLL} to produce good uncertainty estimates in the binary classification setting in Appendix Section~\ref{appendix:consistency}.
\end{itemize}

\begin{figure}[b!]
    \centering
    \vspace{-1em}
    \includegraphics[width=\linewidth]{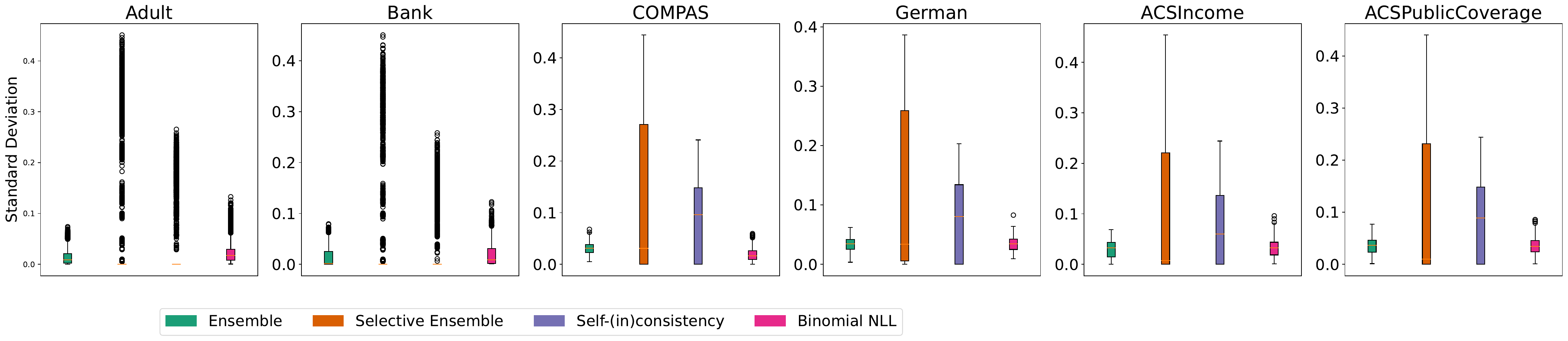}
    \vspace{-1em}
    \caption{This boxplot shows the standard deviation of each individual's uncertainty estimates across different \textit{max\_depth} hyperparameter settings. For example, if an individual has the same uncertainty estimate for each hyper-parameter setting, then their standard deviation is 0 (perfect \textit{consistency}) whereas if they vary wildly, the standard deviation is high (not \textit{consistent}). The \textit{Binomial NLL} and \textit{Ensemble} methods exhibits are the most \textit{consistent}.}
    \label{fig:binary_consistency}
\end{figure}

\textbf{Consistency}~~~Our first experiment evaluates consistency between similar learning pipelines. 
We do this by varying a hyperparameter setting for our model class; ideally, we would select parameters that affect how the model makes predictions, but shouldn't substantially affect the error rate of those predictions. We note that \FU makes it easy to experiment with many hyperparameter settings.
In this paper, we first experiment with the XGBoost hyperparameter \textit{max\_depth}; \textit{depth} is widely applicable to many kinds of models and there are often many equally-valid settings (\FU ensures that all depths produce models with similar accuracy).
The second is \textit{reduction\_threshold} $\gamma$, which is specific to the XGBoost model, and smoothly interpolates between encouraging more or less complex models through tree splits.
Figure \ref{fig:binary_consistency} plots heteroscedastic uncertainty estimates for all individuals in each dataset for \textit{max\_depth} (see Figure~\ref{fig:binary_consistency_gamma} for the \textit{reduction\_threshold} plot, and see Figure~\ref{fig:consistency_range} for a more granular view where we plot the consistency ranges for \textit{each individual} in the data, both in the Appendix).
For some individuals, all the methods consistently output the same uncertainty estimates but, for others, estimates produced by the \textit{Selective Ensemble}  and \textit{Self-(in)consistency Ensemble} vary drastically.
We measure the overall \emph{consistency} of each method by computing the empirical standard deviation of the estimates at each depth.
Table~\ref{tab:binary} reports the maximum individual empirical standard deviation for each algorithm and dataset.
The \textit{Ensemble} algorithm is the most consistent, followed closely by \textit{Binomial NLL}.

\begin{table}[t!]
    \centering
    \caption{Comparison of \emph{calibration} and \emph{consistency} for each algorithm on each binary dataset. Here, calibration is measured by negative log-likelihood while the consistency is measured by the maximum individual empirical standard deviation. The $\pm$ indicates the standard deviation of these values over 10 iterations. Note that we adopt the Olympic medal convention in all Tables throughout our paper: \colorbox{gold!30}{gold}, \colorbox{silver!30}{silver} and \colorbox{bronze!30}{bronze} cells signify first, second and third best performance, respectively.}
    \vspace{-0.6em}
    \label{tab:binary}
    \resizebox{\linewidth}{!}{\begin{tabular} {lccccc||cccccc}
    \toprule
    & \multicolumn{5}{c}{\textbf{Calibration (Negative Log-Likelihood)}} & \multicolumn{5}{c}{\textbf{Consistency}} \\ 
    \textbf{Approach} & \textbf{ACS} & \textbf{Adult} & \textbf{Bank} & \textbf{COMPAS} & \textbf{German} & \textbf{ACS} & \textbf{Adult} & \textbf{Bank} & \textbf{COMPAS} & \textbf{German} \\ \midrule
    \textit{Ensemble} & \cellcolor{bronze!30}\footnotesize{1.1 \smallerpm 0.04} & \footnotesize{0.9 \smallerpm 0.02} & \cellcolor{bronze!30}\footnotesize{0.5 \smallerpm 0.02} & \cellcolor{bronze!30}\footnotesize{1.8 \smallerpm 0.07} & \cellcolor{bronze!30}\footnotesize{0.97 \smallerpm 0.12} & \cellcolor{gold!30}\footnotesize{0.08 \smallerpm 0.00} & \cellcolor{gold!30}\footnotesize{0.08 \smallerpm 0.01} & \cellcolor{gold!30}\footnotesize{0.09 \smallerpm 0.01} & \cellcolor{gold!30}\footnotesize{0.06 \smallerpm 0.00} & \cellcolor{gold!30}\footnotesize{0.062 \smallerpm 0.00} \\ 
    \emph{Selective Ens.} & \cellcolor{bronze!30}\footnotesize{1.1 \smallerpm 0.05} & \cellcolor{bronze!30}\footnotesize{0.88 \smallerpm 0.02} & \cellcolor{bronze!30}\footnotesize{0.5 \smallerpm 0.02} & \cellcolor{bronze!30}\footnotesize{1.8 \smallerpm 0.08} & \footnotesize{1.0 \smallerpm 0.16} & \footnotesize{0.45 \smallerpm 0.01} & \footnotesize{0.45 \smallerpm 0.01} & \footnotesize{0.45 \smallerpm 0.01} & \footnotesize{0.44 \smallerpm 0.01} & \footnotesize{0.40 \smallerpm 0.02} \\ 
    \textit{(In)cons. Ens.} & \cellcolor{silver!30}\footnotesize{1.0 \smallerpm 0.04} & \cellcolor{silver!30}\footnotesize{0.82 \smallerpm 0.03} & \cellcolor{silver!30}\footnotesize{0.42 \smallerpm 0.02} & \cellcolor{silver!30}\footnotesize{1.5 \smallerpm 0.07} & \cellcolor{silver!30}\footnotesize{0.82 \smallerpm 0.13} & \cellcolor{bronze!30}\footnotesize{0.26 \smallerpm 0.01} & \cellcolor{bronze!30}\footnotesize{0.26 \smallerpm 0.01} & \cellcolor{bronze!30}\footnotesize{0.25 \smallerpm 0.01} & \cellcolor{bronze!30}\footnotesize{0.25 \smallerpm 0.01} & \cellcolor{bronze!30}\footnotesize{0.21\smallerpm 0.01} \\ 
    \textit{Binom. NLL} & \cellcolor{gold!30}\footnotesize{0.4 \smallerpm 0.01} & \cellcolor{gold!30}\footnotesize{0.31 \smallerpm 0.0} & \cellcolor{gold!30}\footnotesize{0.2 \smallerpm 0.0} & \cellcolor{gold!30}\footnotesize{0.6 \smallerpm 0.01} & \cellcolor{gold!30}\footnotesize{0.5 \smallerpm 0.04} & \cellcolor{silver!30}\footnotesize{0.10 \smallerpm 0.01} & \cellcolor{silver!30}\footnotesize{0.13 \smallerpm 0.01} & \cellcolor{silver!30}\footnotesize{0.12 \smallerpm 0.01} & \cellcolor{silver!30}\footnotesize{0.07 \smallerpm 0.00} & \cellcolor{silver!30}\footnotesize{0.08 \smallerpm 0.01} \\ 
    \bottomrule
    \end{tabular}}
    \vspace{-1.5em}
\end{table}

\begin{figure}[b]
    \centering
    \vspace{-1.5em}
    \includegraphics[width=\linewidth]{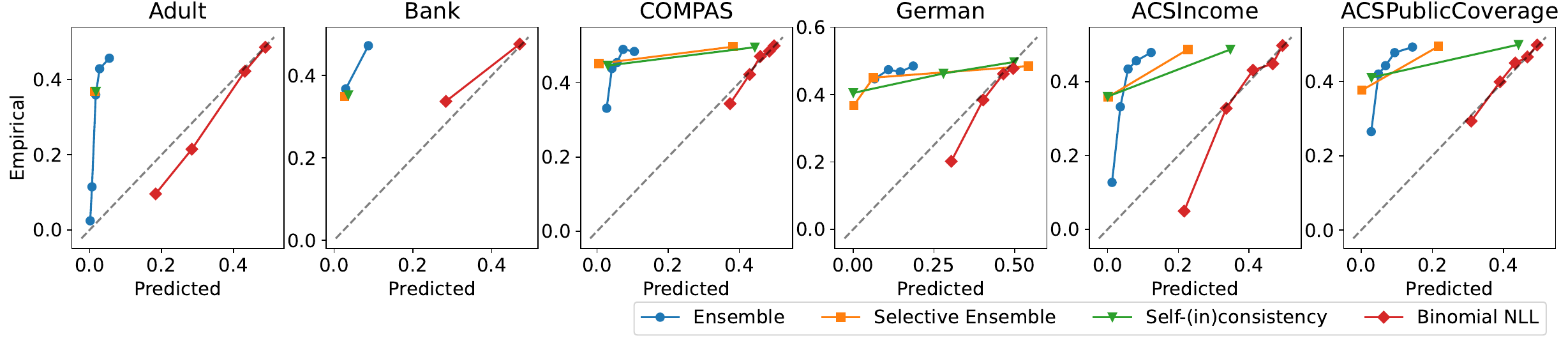}
    \vspace{-2em}
    \caption{
    For five groups assembled by predicted uncertainty, we plot the average predicted uncertainty against the empirical standard deviation of the outcomes. An algorithm is perfectly calibrated if predicted uncertainty equals the empirical standard deviation i.e., the points lie on the dashed identity line. Note that uncertainty estimates do not always represent variance, so we expect a positive but not necessarily linear correlation. Additionally, note that this \textit{calibration} graph also reflects \textit{consistency}; a less \textit{consistent} method will have a more arbitrary grouping leading to a flatter observed slope.}
    \label{fig:binary_calibration}
\end{figure}

\textbf{Calibration (qualitative)}~~~The next experiment qualitatively evaluates calibration.
For each method, we identify groups of individuals with similar uncertainty estimates and empirically evaluate the standard deviation of the residual difference between the observed and predicted outcomes.
Results in Figure~\ref{fig:binary_calibration} demonstrate that predicted uncertainty estimates from the \textit{Selective Ensemble} and \textit{Self-(in)consistency} algorithms do not appear related to the empirical standard deviation.
In contrast, the estimates from the \textit{Ensemble} and \textit{Binomial NLL} methods are clearly related to the empirical standard deviation; we note that the \textit{Binomial NLL} method most closely tracks the identity line.

\textbf{Calibration (quantitative)}~~~Finally, we quantitatively evaluate calibration by interpreting uncertainty estimates as the parameter of a fixed distribution.
In the binary setting, the distribution is completely described by a single probability and the Binomial distribution.
Table~\ref{tab:binary} gives the negative log-likelihood for each method on each dataset. To produce Table~\ref{tab:binary}, we interpret estimates from models that produce a different kind of uncertainty as a standard deviation for comparison.
By design, the \textit{Binomial NLL} method minimizes the negative log-likelihood on the training set, so unsurprisingly we find that the \textit{Binomial NLL} method also gives the best performance on the test set. Overall, Figures~\ref{fig:binary_consistency} and~\ref{fig:binary_calibration} and Table~\ref{tab:binary} suggest that the \textit{Binomial NLL} method produces heteroscedastic uncertainty estimates that are simultaneously the most \emph{consistent} and \emph{calibrated}.

\subsection{Regression Tasks}\label{sec:cc_regression}
In this section, we turn our attention to evaluating methods for estimating heteroscedastic uncertainty in the regression setting over continuous outcomes. \FU contains several algorithms drawn from the heteroscedastic uncertainty literature.

\begin{itemize}[leftmargin=*]
    \item The \textit{Normal NLL} method learns parameters $\mu$ and $\sigma$ that minimize the negative log-likelihood loss of the normal distribution, given by $\ell(y, \mu, \sigma) =  - \log \sigma + \frac12 \left(\frac{x-\mu}{\sigma} \right)^2$. 
    \item The \textit{$\beta$-NLL} method \citep{seitzer2022pitfalls}  learns a version of the NLL loss that is multiplied by the constant $\sigma^{2\beta}$ (in our experiments we set $\beta=\frac12$ as suggested in the original paper).
    \item The \textit{Faithful NLL} method \citep{stirn2023faithful} learns mean predictions $\mu$ with the standard mean squared error loss while the standard deviation predictions $\sigma$ are learned with the NLL loss.
    \item A natural \textit{Ensemble} method serves as a point of comparison for these heteroscedastic algorithms: \textit{Ensemble} learns an ensemble of models trained on bagged samples of the training set, and outputs the mean and standard deviation over model predictions.
\end{itemize}

\begin{table}[t]
    \centering
    \caption{Comparison of Negative Log-Likelihood (NLL) and Consistency for each algorithm on each dataset. We measure \emph{consistency} as the maximum over individuals of the standard deviation of the predictions they receive with respect to the depth of the model.}
    \vspace{-0.6em}
    \label{tab:regression}
    \resizebox{\linewidth}{!}{\begin{tabular} {lccccc||ccccc}
    \toprule
    & \multicolumn{5}{c}{\textbf{Calibration (Negative Log-Likelihood)}} & \multicolumn{5}{c}{\textbf{Consistency}} \\ 
    \textbf{Approach} & \textbf{Comm.} & \textbf{IHDP} & \textbf{Insur.} & \textbf{Law} & \textbf{Twins} & \textbf{Comm.} & \textbf{IHDP} & \textbf{Insur.} & \textbf{Law} & \textbf{Twins} \\ \midrule
    \textit{Ensemble} & 120.0 $\pm$ 21.0 & 3.9 $\pm$ 1.3 & 21.0 $\pm$ 15.0 & 200.0 $\pm$ 21.0 & 160.0 $\pm$ 9.3 & \cellcolor{gold!30}\footnotesize{0.05 \smallerpm 0.01} & \cellcolor{gold!30}\footnotesize{0.02 \smallerpm 0.00} & \cellcolor{gold!30}\footnotesize{0.03 \smallerpm 0.00} & \cellcolor{gold!30}\footnotesize{0.05 \smallerpm 0.01} & \cellcolor{gold!30}\footnotesize{0.09 \smallerpm 0.01} \\ 
    \textit{Normal NLL} & \cellcolor{bronze!30}0.35 $\pm$ 0.08 & \cellcolor{gold!30}-0.6 $\pm$ 0.07 & \cellcolor{gold!30}-0.67 $\pm$ 0.04 & \cellcolor{gold!30}-0.25 $\pm$ 0.0 & \cellcolor{gold!30}0.41 $\pm$ 0.0 & \footnotesize{0.12 \smallerpm 0.01} & \footnotesize{0.08 \smallerpm 0.01} & \footnotesize{0.09 \smallerpm 0.01} & \footnotesize{0.11 \smallerpm 0.01} & \cellcolor{bronze!30}\footnotesize{0.20 \smallerpm 0.01} \\ 
    \textit{$\beta$-NLL} & \cellcolor{gold!30}0.34 $\pm$ 0.08 & \cellcolor{gold!30}-0.6 $\pm$ 0.02 & \cellcolor{silver!30}-0.65 $\pm$ 0.03 & \cellcolor{gold!30}-0.25 $\pm$ 0.0 & \cellcolor{gold!30}0.41 $\pm$ 0.0 & \cellcolor{silver!30}\footnotesize{0.11 \smallerpm 0.01} & \cellcolor{silver!30}\footnotesize{0.06 \smallerpm 0.01} & \cellcolor{silver!30}\footnotesize{0.06 \smallerpm 0.01} & \cellcolor{silver!30}\footnotesize{0.08 \smallerpm 0.00} & \footnotesize{3.26 \smallerpm 7.66} \\ 
    \textit{Faithful NLL} & \cellcolor{gold!30}0.34 $\pm$ 0.11 & \cellcolor{bronze!30}-0.58 $\pm$ 0.01 & \cellcolor{bronze!30}-0.64 $\pm$ 0.03 & \cellcolor{gold!30}-0.25 $\pm$ 0.0 & \cellcolor{gold!30}0.41 $\pm$ 0.0 & \cellcolor{bronze!30}\footnotesize{0.12 \smallerpm 0.01} & \cellcolor{bronze!30}\footnotesize{0.08 \smallerpm 0.01} & \cellcolor{bronze!30}\footnotesize{0.07 \smallerpm 0.01} & \cellcolor{bronze!30}\footnotesize{0.09 \smallerpm 0.01} & \cellcolor{silver!30}\footnotesize{0.19 \smallerpm 0.01} \\ 
    \bottomrule
    \end{tabular}}
    \vspace{-1em}
\end{table}

Using \FU, we evaluate these methods for \emph{consistency} and \emph{calibration}.
We evaluate \emph{consistency} in the same manner in the regression setting as we did in the binary setting, varying two hyperparameters while checking that the maximum deviation of uncertainty estimates remains stable (see Figure~\ref{fig:regression_consistency} in the Appendix). In Table~\ref{tab:regression}, we see that the \textit{Ensemble} approach is the most \emph{consistent}, although for four of the five datasets, \textit{$\beta$-NLL} performs similarly.

\emph{Calibration} is also evaluated in the same manner for regression tasks as for binary. Figure~\ref{fig:regression_calibration} in the Appendix provides the qualitative assessment, while quantitatively Table \ref{tab:regression} gives the NLL for each algorithm on each dataset.
Our results demonstrate that though the \textit{Ensemble} method is \emph{consistent}, its predictions are very poorly \emph{calibrated}. In contrast, all NLL approaches achieve a similar level of \emph{calibration} across all five regression datasets. Overall, the various NLL approaches are all comparably \emph{consistent} to the \textit{Ensemble} method and are significantly more \emph{calibrated}.

\section{Abstaining on Classification Tasks}\label{sec:abstaining}

We now turn our attention to incorporating uncertainty estimates into \textit{fair} algorithms for classification, to assess their impact on downstream fairness metrics, focusing on the abstention framework where models are allowed to \textit{abstain} from making a prediction if the heteroscedastic uncertainty estimates are too large \citep{black2021selective,cooper2024arbitrariness}. \FU instantiates this intervention, allowing models to abstain from predicting under high uncertainty. We find that while the abstention framework allows models to abstain from incorrect predictions, unsurprisingly reducing the overall error rate, it does not make the distribution of predictions more balanced between demographic groups. Moreover, the flexibility afforded by abstention is also its biggest limitation; if allowed to make almost no predictions, a model can easily achieve optimal performance without meaningful outputs for the majority of observations. Thus, \FU crucially focuses on the question of abstention \textit{rate} and how it affects overall error, statistical parity, equalized odds, etc.

\begin{figure}[b!]
    \centering
    \vspace{-1.5em}
    \includegraphics[width=\linewidth]{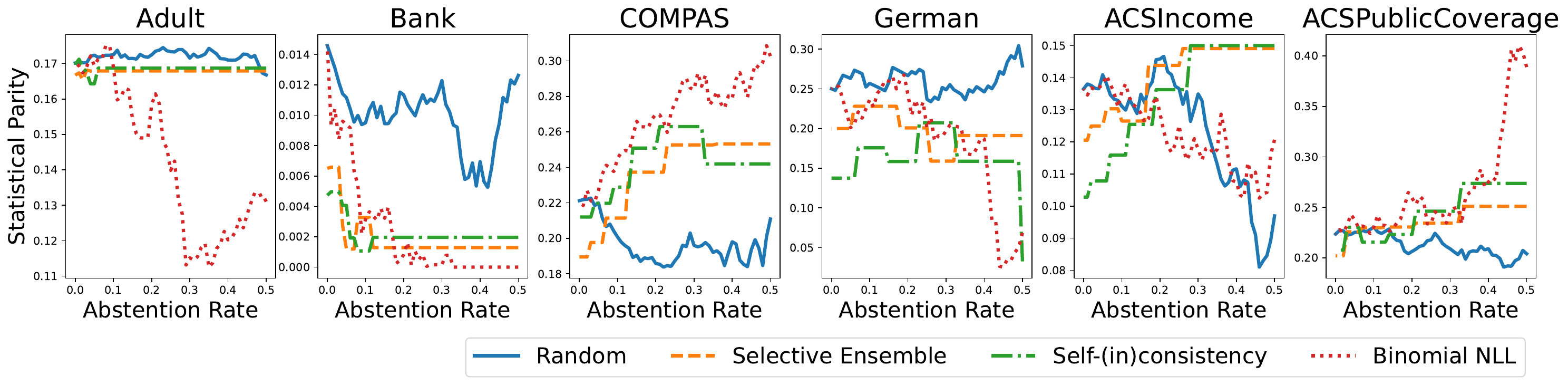}
    \vspace{-2em}
    \caption{Abstaining has no reliable effect on \textit{Statistical Parity} (comparable to the \textit{Random} baseline).} 
    \label{fig:binary_abstention}
\end{figure}

\begin{figure}[h]
    \centering
    \includegraphics[width=\linewidth]{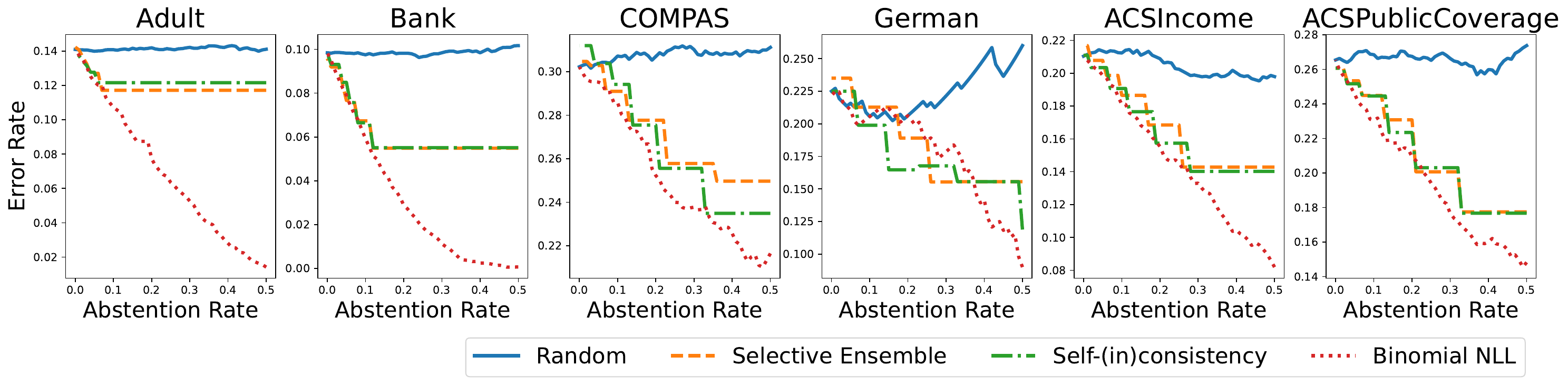}
    \vspace{-1.5em}
    \caption{For an abstention rate $r$, \FU abstains on the $r$ fraction of observations with the highest uncertainty. For heteroscedastic uncertainty methods, predictions become more accurate as the model abstains more, while the error rate for the random baseline remains steady.}
    \label{fig:binary_abstention_er}
    \vspace{-1em}
\end{figure}

In Figure \ref{fig:binary_abstention_er}, we see that allowing models to make fewer predictions based on uncertainty estimates decreases their error. In contrast, a \textit{Random} baseline (real binary predictions but \textit{random} uncertainty estimates) maintains the same error rate. Notice that the \textit{Selective Ensemble} and \textit{Self-(in)consistency Ensemble} produce the same uncertainty estimates for many observations, hence the flat lines. While abstaining invariably reduces overall error, it has a more chaotic impact on \textit{Statistical Parity}. Figure \ref{fig:binary_abstention} shows that abstaining can improve (decrease) or worsen (increase) statistical parity unpredictably across different data distributions; here, the model behavior resembles the \textit{Random} uncertainty prediction baseline. We observe a similar trend in Figure~\ref{fig:abstention_eo}, which plots the \textit{Equalized Odds} fairness metric \citep{hardt2016equality} against the abstention rate (see Appendix~\ref{appendix:abstain}).

\begin{table}[t!]
    \centering
    \caption{Various evaluation and fairness metrics for each algorithm on the ACS Income (see Appendix~\ref{app:fairness_metrics} for definitions). While the abstention framework allows models to reduce their \textit{Error Rate}, it does \textit{not} magically reduce the imbalance in outcomes between demographic groups.}
    \vspace{-0.6em}
    \label{tab:fairness_acs}
    \resizebox{\textwidth}{!}{\begin{tabular} {lcccccccc}
\toprule
\textbf{Approach} & \textbf{Error Rate} & \textbf{Stat. Parity} & \textbf{Equalized Odds} & \textbf{Equal Opp.} & \textbf{Disp. Impact} & \textbf{Predictive Parity} & \textbf{FPR} & \textbf{Included \%} \\ \midrule
\textit{Baseline} & 0.22 $\pm$ 0.01 & 0.18 $\pm$ 0.02 & 0.17 $\pm$ 0.05 & 0.17 $\pm$ 0.05 & 1.7 $\pm$ 0.14 & 0.057 $\pm$ 0.03 & 0.083 $\pm$ 0.02 & \cellcolor{gold!30}100 $\pm$ 0.0 \\ 
\textit{Threshold Optimizer SP} & 0.23 $\pm$ 0.01 & \cellcolor{gold!30}0.029 $\pm$ 0.02 & 0.14 $\pm$ 0.05 & \cellcolor{gold!30}0.063 $\pm$ 0.03 & \cellcolor{silver!30}0.98 $\pm$ 0.07 & 0.24 $\pm$ 0.04 & 0.13 $\pm$ 0.05 & \cellcolor{gold!30}100 $\pm$ 0.0 \\ 
\textit{Threshold Optimizer EO} & 0.23 $\pm$ 0.01 & 0.10 $\pm$ 0.03 & \cellcolor{gold!30}0.080 $\pm$ 0.05 & \cellcolor{bronze!30}0.079 $\pm$ 0.05 & 1.3 $\pm$ 0.11 & 0.12 $\pm$ 0.04 & \cellcolor{gold!30}0.024 $\pm$ 0.02 & \cellcolor{gold!30}100 $\pm$ 0.0 \\ 
\textit{Exponentiated Gradient SP} & 0.23 $\pm$ 0.01 & \cellcolor{silver!30}0.041 $\pm$ 0.02 & \cellcolor{bronze!30}0.12 $\pm$ 0.04 & \cellcolor{silver!30}0.065 $\pm$ 0.04 & \cellcolor{gold!30}0.95 $\pm$ 0.08 & 0.20 $\pm$ 0.03 & 0.12 $\pm$ 0.04 & \cellcolor{gold!30}100 $\pm$ 0.0 \\ 
\textit{Exponentiated Gradient EO} & 0.22 $\pm$ 0.01 & 0.10 $\pm$ 0.04 & \cellcolor{silver!30}0.092 $\pm$ 0.05 & 0.090 $\pm$ 0.05 & 1.3 $\pm$ 0.15 & 0.12 $\pm$ 0.05 & \cellcolor{silver!30}0.031 $\pm$ 0.02 & \cellcolor{gold!30}100 $\pm$ 0.0 \\ 
\textit{Grid Search SP} & 0.23 $\pm$ 0.01 & \cellcolor{bronze!30}0.083 $\pm$ 0.05 & 0.14 $\pm$ 0.06 & 0.11 $\pm$ 0.07 & \cellcolor{bronze!30}1.0 $\pm$ 0.26 & 0.18 $\pm$ 0.05 & 0.11 $\pm$ 0.07 & \cellcolor{gold!30}100 $\pm$ 0.0 \\ 
\textit{Grid Search EO} & 0.23 $\pm$ 0.01 & 0.15 $\pm$ 0.03 & 0.13 $\pm$ 0.07 & 0.13 $\pm$ 0.07 & 1.5 $\pm$ 0.15 & 0.087 $\pm$ 0.05 & \cellcolor{bronze!30}0.056 $\pm$ 0.03 & \cellcolor{gold!30}100 $\pm$ 0.0 \\ 
\midrule
\textit{Random} & 0.22 $\pm$ 0.01 & 0.17 $\pm$ 0.03 & 0.15 $\pm$ 0.05 & 0.15 $\pm$ 0.05 & 1.6 $\pm$ 0.16 & 0.044 $\pm$ 0.03 & 0.085 $\pm$ 0.02 & 88 $\pm$ 7.0 \\ 
\textit{Ensemble} & \cellcolor{bronze!30}0.20 $\pm$ 0.02 & 0.19 $\pm$ 0.03 & 0.16 $\pm$ 0.06 & 0.16 $\pm$ 0.06 & 1.7 $\pm$ 0.18 & \cellcolor{gold!30}0.032 $\pm$ 0.04 & 0.093 $\pm$ 0.03 & 89 $\pm$ 8.0 \\ 
\textit{Selective Ensemble} & 0.20 $\pm$ 0.02 & 0.18 $\pm$ 0.03 & 0.15 $\pm$ 0.05 & 0.15 $\pm$ 0.06 & 1.7 $\pm$ 0.20 & 0.039 $\pm$ 0.04 & 0.082 $\pm$ 0.03 & 94 $\pm$ 7.3 \\ 
\textit{Self-(in)consistency} & \cellcolor{silver!30}0.19 $\pm$ 0.03 & 0.19 $\pm$ 0.03 & 0.16 $\pm$ 0.05 & 0.15 $\pm$ 0.06 & 1.7 $\pm$ 0.20 & \cellcolor{bronze!30}0.035 $\pm$ 0.03 & 0.081 $\pm$ 0.03 & 89 $\pm$ 8.9 \\ 
\textit{Binomial NLL} & \cellcolor{gold!30}0.18 $\pm$ 0.03 & 0.19 $\pm$ 0.02 & 0.15 $\pm$ 0.04 & 0.15 $\pm$ 0.05 & 1.8 $\pm$ 0.14 & \cellcolor{silver!30}0.034 $\pm$ 0.03 & 0.080 $\pm$ 0.02 & 83 $\pm$ 9.5 \\ 
\bottomrule
\end{tabular}
}
    \vspace{-1.5em}
\end{table}

Our benchmark additionally evaluates uncertainty methods against standard baseline and state-of-the-art fairness algorithms. To construct this comparison, \FU tests the \textit{Random}, \textit{Selective Ensemble}, \textit{Self-(in)consistency}, and \textit{Binomial NLL} methods at inclusion rates between $75\%$ and $100\%$. It selects an abstention rate for comparison by optimizing a simple objective function that is the normalized sum of the \textit{Error Rate}, \textit{Statistical Parity}, and \textit{Equalized Odds}. Table~\ref{tab:fairness_acs} gives performances on the ACS Income task \citep{ding2021retiring}  (results for the other datasets appear in Appendix \ref{appendix:binary_fair}).
Once again, methods allowed to abstain had lower error (\textit{Self-(in)consistency} and \textit{Binomial NLL} methods had the lowest). However, across 5 of the 6 fairness metrics reported, methods that did not abstain were the \textit{highest} performing in terms of fairness. This is surprising, as it seems directly opposed to one of the main claims in \cite{cooper2024arbitrariness}.

\begin{figure}[b]
    \centering
    \vspace{-1em}
    \includegraphics[width=0.9\linewidth]{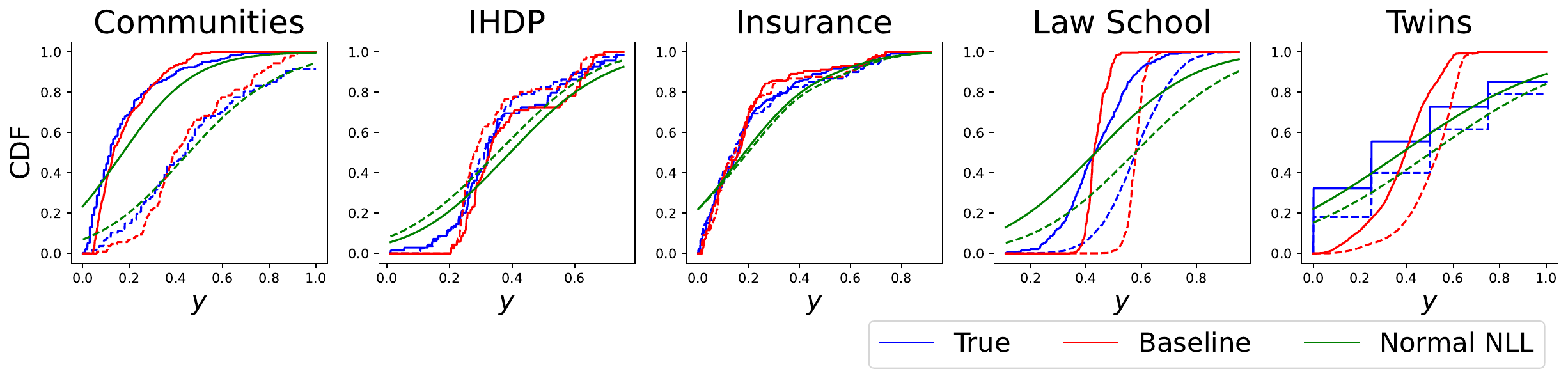}
    \vspace{-1em}
    \caption{CDF by group of on each dataset. The solid lines indicate one protected group and the dashed lines indicate the other protected group. Normal NLL effectively smooths the CDF.}
    \label{fig:regression_fairness}
\end{figure}

A natural question arises given the results in Table~\ref{tab:fairness_acs}: for which individuals do the abstaining models choose not to predict? In Figure~\ref{fig:wasserstein} (deferred to Appendix~\ref{appendix:abstain}), we address this by comparing the empirical distributions of the overall population and the included (non-abstained) population for each feature across methods. We compute the average Wasserstein distance between these distributions (averaged over methods) and plot the feature with the largest distance. The variables with the greatest differences tend to be protected attributes like marriage status and sex. However, except for the \textit{Binomial NLL} distribution on the \textit{Adult} dataset, these differences are small between the overall distribution and the one selected through the abstaining process.

\section{Uncertainty Aware Fair Regression}\label{sec:fairness_regression}

So far, we have used \FU to demonstrate that incorporating heteroscedastic uncertainty estimates into \textit{fair} classification algorithms requires care; this is no different in the regression setting. By outputting both predictions and uncertainty estimates in either setting, downstream users have the flexibility to apply uncertainty estimates in domain-appropriate ways \textit{beyond abstention}. For example, practitioners can flag high uncertainty estimates for manual review \citep{madras2018predict}, combine predictions and uncertainty estimates into domain-specific metrics \citep{ali2021accounting}, or discover patterns of high uncertainty that indicate a lack of meaningful features, prompting more data collection \citep{chen2018my}. 
Still, it remains an open question how to measure the fairness for models that output \textit{both} predictions and uncertainty estimates, particularly for regression models. 
To address this, we propose a new fairness metric (included in \FU), \textit{uncertainty-aware statistical parity (\textit{UA-SP})} (Definition~\ref{def:uasp}), which naturally incorporates the available uncertainty estimates into the the standard regression-specific form of \textit{Statistical Parity} (Definition~\ref{def:sp}).

Note that \textit{UA-SP} (Definition~\ref{def:uasp}) applies to models which produce point predictions $\mu$ \textit{and} uncertainty estimates $\sigma$ for each observation. These estimates are treated as model-derived inputs to our fairness metric, not as the underlying true parameters. We model predictions as distributions to incorporate the inherent uncertainty of model outputs; this approach is particularly salient in settings with heteroscedastic variance. When the uncertainty estimates are zero, our metric reduces to the standard statistical parity definition, thus generalizing it to account for variability in predictions due to uncertainty. The generalized definition assumes that the uncertainty follows a normal distribution with mean $\mu$ and standard deviation $\sigma$. The validity of this assumption depends on the setting.

\begin{definition}[Statistical Parity]\label{def:sp}
Consider a method $f: \mathcal{X} \times \mathcal{A} \to \mathcal{Y}$ that outputs a mean prediction.
Then $f$ satisfies statistical parity if~
$\Pr(f(\mathbf{X},A) \geq y | A=a)
= \Pr(f(\mathbf{X}, A) \geq y)$
for all protected groups $a \in \mathcal{A}$ and outcomes $y \in \mathcal{Y}$.
\end{definition}

\begin{definition}[Uncertainty-Aware Statistical Parity (\textit{UA-SP})]\label{def:uasp}
Consider $f: \mathcal{X} \times \mathcal{A} \to \mathcal{Y} \times \mathbb{R}$ that estimates a mean $\mu$ and a standard deviation $\sigma$.
The predictions induce a randomized function $\tilde{f}: \mathcal{X} \times \mathcal{A} \to \mathcal{Y}$ that samples $y \sim \mathcal{N}(\mu,\sigma^2)$. Then $f$ satisfies uncertainty-aware statistical parity if $~\Pr(\tilde{f}(\mathbf{X},A) \geq y | A=a)
= \Pr(\tilde{f}(\mathbf{X}, A) \geq y)$
for all protected groups $a \in \mathcal{A}$ and outcomes $y \in \mathcal{Y}$.
\end{definition}

Definition \ref{def:uasp} incorporates uncertainty estimates in a natural way: the predicted distribution is smoothed for observations with high uncertainty.
In contrast, Definition \ref{def:sp} holds the algorithms to a stringent standard even when heteroscedastic variance is large.

In regression settings, Definition~\ref{def:sp} is often interpreted in terms of a Cumulative Density Function (CDF) over predictions \citep{agarwal2019fair, liu2022conformalized}; uncertainty-aware statistical parity (Definition~\ref{def:uasp}) shares this interpretation. For methods that only produce predictions, the CDF is with respect to the randomness of the observations $\mathbf{x}$.
For methods that produce predictions \textit{and} uncertainty estimates, the CDF is with respect to the randomness of the observations and the induced randomized function $\tilde{f}$. Figure~\ref{fig:regression_fairness} gives CDFs for \textit{True}, \textit{Baseline} and \textit{Normall NLL}: here, \textit{True} indicates the correct labels in the test data, \textit{Baseline} indicates a standard regression model without any fairness intervention, and \textit{Normal NLL} is the smoothed method as described in Section~\ref{sec:cc_regression} (for visual clarity, we present comparisons only against \textit{Normal NLL}, which is similar to \textit{Faithful NLL} and more \emph{consistent} and \emph{calibrated} than \textit{Ensemble} and \textit{$\beta$-NLL}).

Figure \ref{fig:regression_fairness} illustrates how incorporating uncertainty into model predictions effectively smooths the CDF. In the case of the Twins dataset, we observe that this smoothing is both strong and accurate, allowing the model to closely capture the true distribution of the labels. Table \ref{tab:regression_fairness} presents fairness metrics computed according to either Definition~\ref{def:sp} or Definition~\ref{def:uasp}, depending on whether the model provides uncertainty estimates. The \textit{Normal NLL} method significantly outperforms others in terms of fairness, even surpassing the \textit{Ensemble} method. Remarkably, the \textit{Normal NLL} method, which is both competitively \emph{consistent} and \emph{calibrated}, achieves substantial fairness improvements \textit{without} any explicit fairness interventions.

\begin{table}[t]
    \centering
    \caption{Measuring (uncertainy-aware) statistical parity (\textit{UA-SP}) via Kolmogorov-Smirnov distance: the maximum over outputs $y$ of the distance between per-group CDFs \citep{agarwal2019fair}. Simply using \emph{consistent} and \emph{calibrated} uncertainty methods substantially reduces \textit{UA-SP}.}
    \vspace{-0.6em}
    \label{tab:regression_fairness}
    \resizebox{1\linewidth}{!}{\begin{tabular} {lccccc}
\toprule
\textbf{Approach} & \textbf{Communities} & \textbf{IHDP} & \textbf{Insurance} & \textbf{Law School} & \textbf{Twins} \\ \midrule
True & 0.561 $\pm$ 0.036 & 0.153 $\pm$ 0.061 & 0.1 $\pm$ 0.022 & 0.484 $\pm$ 0.021 & 0.161 $\pm$ 0.011 \\ 
Baseline & 0.644 $\pm$ 0.025 & 0.144 $\pm$ 0.05 & 0.091 $\pm$ 0.019 & 0.965 $\pm$ 0.009 & 0.404 $\pm$ 0.009 \\ 
Exponentiated Gradient Square & 0.638 $\pm$ 0.024 & 0.16 $\pm$ 0.053 & 0.094 $\pm$ 0.025 & 0.902 $\pm$ 0.015 & 0.357 $\pm$ 0.01 \\ 
Exponentiated Gradient Absolute & 0.638 $\pm$ 0.024 & 0.157 $\pm$ 0.057 & 0.094 $\pm$ 0.025 & 0.902 $\pm$ 0.015 & 0.359 $\pm$ 0.009 \\ 
Grid Search Square & 0.649 $\pm$ 0.035 & 0.167 $\pm$ 0.056 & 0.093 $\pm$ 0.025 & 0.888 $\pm$ 0.019 & 0.357 $\pm$ 0.007 \\ 
Grid Search Absolute & 0.649 $\pm$ 0.035 & 0.167 $\pm$ 0.056 & 0.092 $\pm$ 0.024 & 0.888 $\pm$ 0.019 & 0.358 $\pm$ 0.007 \\ 
\midrule
Ensemble & 0.62 $\pm$ 0.021 & 0.101 $\pm$ 0.052 & 0.076 $\pm$ 0.022 & 0.939 $\pm$ 0.008 & 0.389 $\pm$ 0.009 \\ 
Normal NLL & \cellcolor{silver!30}0.37 $\pm$ 0.016 & \cellcolor{bronze!30}0.048 $\pm$ 0.028 & \cellcolor{silver!30}0.052 $\pm$ 0.022 & \cellcolor{bronze!30}0.196 $\pm$ 0.003 & \cellcolor{silver!30}0.096 $\pm$ 0.002 \\ 
$\beta$-NLL & \cellcolor{bronze!30}0.375 $\pm$ 0.019 & \cellcolor{silver!30}0.045 $\pm$ 0.025 & \cellcolor{gold!30}0.051 $\pm$ 0.022 & \cellcolor{silver!30}0.195 $\pm$ 0.004 & \cellcolor{silver!30}0.096 $\pm$ 0.003 \\ 
Faithful NLL & \cellcolor{gold!30}0.299 $\pm$ 0.151 & \cellcolor{gold!30}0.042 $\pm$ 0.021 & \cellcolor{bronze!30}0.054 $\pm$ 0.02 & \cellcolor{gold!30}0.192 $\pm$ 0.004 & \cellcolor{gold!30}0.095 $\pm$ 0.003 \\ 
\bottomrule
\end{tabular}
}
\end{table}

\subsection{Related Work}

Prior work in algorithmic fairness largely focuses on fixed fairness constraints, framing the issue as optimizing along a Pareto frontier between accuracy and fairness metrics that may conflict \citep{hardt2016equality, zafar2017fairness, mitchell2021algorithmic, bell2023possibility, wang2024aleatoric}. More recent research has introduced \textit{uncertainty} estimation as a crucial factor in model selection under fairness. Particularly relevant are \citet{black2021selective} and \citet{cooper2024arbitrariness}, who explore ensemble approaches to estimate prediction uncertainty based on the standard deviation from an ensemble of models. These methods, as we have demonstrated, tend to focus on artifacts of the learning pipeline rather than inherent heteroscedastic uncertainty.

Other studies examine uncertainty estimation through model multiplicity \citep{black2022model}, analyze the relationship between inadequate sample sizes and disparate epistemic uncertainty among subgroups \citep{chen2018my, zhang2021assessing}, and extend abstention frameworks to regression \citep{shah2022selective}. Additionally, tailored inference and prediction models under heteroscedastic assumptions have been extensively studied in economics and statistics, particularly with Bayesian inference \citep{white1980heteroskedasticity,mackinnon2012thirty,rigobon2003identification,hayes2007using, ji2020can}. The impact of Bayesian conditioning via latent variables on fairness has also been considered as an alternative to bagging for uncertainty estimation, though \emph{calibration} of these estimates can be challenging due to the large space of priors \citep{mcnair2018preventing, ji2020can}. More recently, loss functions accommodating heteroscedastic assumptions in machine learning have improved uncertainty estimation and model robustness \citep{collier2020simple, abdar2021review}, with some work addressing heteroscedastic pitfalls in log-likelihood-based loss functions 
\citep{seitzer2022pitfalls, stirn2023faithful}.

Prior work on uncertainty in fair regression has focused on estimating quantiles -- confidence intervals -- for predictions \citep{liu2022conformalized,kuzucu2023uncertainty,wang2024variational}.
Quantile predictions are considered fair if they are equally calibrated and accurate for different demographic groups \citep{wang2024aleatoric,wang2024equal}.
However, quantiles are only valid for a given probability threshold and fail to adequately describe the distribution. For example, recall how Figure~\ref{fig:distributions} demonstrated how the mean and quantiles could be the same for two distributions that are substantially different.

\section{Conclusion}\label{sec:conclusion}
\FU actualizes the \emph{consistency} and \emph{calibration} axioms in the form of a robust benchmark for evaluating uncertainty estimates in fair predictions. \FU suggests the following results: In the binary setting, natural uncertainty estimates beat complex ensemble based approaches and abstaining improves error but not imbalance between demographic groups. In the regression setting, \emph{consistent} and \emph{calibrated} uncertainty methods can reduce distributional imbalance without any explicit fairness intervention. The current version of \FU is not without limitations: different models \citep{gorishniy2021revisiting}, parameters, and datasets could be varied to assess \emph{consistency}, different metrics introduced for \emph{calibration}, and additional fairness interventions explored. Because of its specialized nature, we expect a net positive social impact from \FU. We hope that the extensible package construction allows \FU to grow with the community.

\section*{Acknowledgements}
Both LR and RTW were supported by the National Science Foundation under Graduate Research Fellowship Grant No. DGE-2234660.

\bibliography{references}

\begin{thebibliography}{60}
\providecommand{\natexlab}[1]{#1}
\providecommand{\url}[1]{\texttt{#1}}
\expandafter\ifx\csname urlstyle\endcsname\relax
  \providecommand{\doi}[1]{doi: #1}\else
  \providecommand{\doi}{doi: \begingroup \urlstyle{rm}\Url}\fi

\bibitem[Abdar et~al.(2021)Abdar, Pourpanah, Hussain, Rezazadegan, Liu,
  Ghavamzadeh, Fieguth, Cao, Khosravi, Acharya, et~al.]{abdar2021review}
Moloud Abdar, Farhad Pourpanah, Sadiq Hussain, Dana Rezazadegan, Li~Liu,
  Mohammad Ghavamzadeh, Paul Fieguth, Xiaochun Cao, Abbas Khosravi, U~Rajendra
  Acharya, et~al.
\newblock A review of uncertainty quantification in deep learning: Techniques,
  applications and challenges.
\newblock \emph{Information fusion}, 76:\penalty0 243--297, 2021.

\bibitem[Agarwal et~al.(2019)Agarwal, Dud{\'\i}k, and Wu]{agarwal2019fair}
Alekh Agarwal, Miroslav Dud{\'\i}k, and Zhiwei~Steven Wu.
\newblock Fair regression: Quantitative definitions and reduction-based
  algorithms.
\newblock In \emph{International Conference on Machine Learning}, pp.\
  120--129. PMLR, 2019.

\bibitem[Ali et~al.(2021)Ali, Lahoti, and Gummadi]{ali2021accounting}
Junaid Ali, Preethi Lahoti, and Krishna~P Gummadi.
\newblock Accounting for model uncertainty in algorithmic discrimination.
\newblock In \emph{Proceedings of the 2021 AAAI/ACM Conference on AI, Ethics,
  and Society}, pp.\  336--345, 2021.

\bibitem[Almond et~al.(2005)Almond, Chay, and Lee]{almond2005costs}
Douglas Almond, Kenneth~Y Chay, and David~S Lee.
\newblock The costs of low birth weight.
\newblock \emph{The Quarterly Journal of Economics}, 120\penalty0 (3):\penalty0
  1031--1083, 2005.

\bibitem[Angwin et~al.(2022)Angwin, Larson, Mattu, and
  Kirchner]{angwin2022machine}
Julia Angwin, Jeff Larson, Surya Mattu, and Lauren Kirchner.
\newblock Machine bias.
\newblock In \emph{Ethics of data and analytics}, pp.\  254--264. Auerbach
  Publications, 2022.

\bibitem[Barocas et~al.(2023)Barocas, Hardt, and
  Narayanan]{barocas2023fairness}
Solon Barocas, Moritz Hardt, and Arvind Narayanan.
\newblock \emph{Fairness and machine learning: Limitations and opportunities}.
\newblock MIT Press, 2023.

\bibitem[Bell et~al.(2023)Bell, Bynum, Drushchak, Zakharchenko, Rosenblatt, and
  Stoyanovich]{bell2023possibility}
Andrew Bell, Lucius Bynum, Nazarii Drushchak, Tetiana Zakharchenko, Lucas
  Rosenblatt, and Julia Stoyanovich.
\newblock The possibility of fairness: Revisiting the impossibility theorem in
  practice.
\newblock In \emph{Proceedings of the 2023 ACM Conference on Fairness,
  Accountability, and Transparency}, pp.\  400--422, 2023.

\bibitem[Bellamy et~al.(2018)Bellamy, Dey, Hind, Hoffman, Houde, Kannan, Lohia,
  Martino, Mehta, Mojsilovic, Nagar, Ramamurthy, Richards, Saha, Sattigeri,
  Singh, Varshney, and Zhang]{aif360-oct-2018}
Rachel K.~E. Bellamy, Kuntal Dey, Michael Hind, Samuel~C. Hoffman, Stephanie
  Houde, Kalapriya Kannan, Pranay Lohia, Jacquelyn Martino, Sameep Mehta,
  Aleksandra Mojsilovic, Seema Nagar, Karthikeyan~Natesan Ramamurthy, John
  Richards, Diptikalyan Saha, Prasanna Sattigeri, Moninder Singh, Kush~R.
  Varshney, and Yunfeng Zhang.
\newblock {AI Fairness} 360: An extensible toolkit for detecting,
  understanding, and mitigating unwanted algorithmic bias, October 2018.
\newblock URL \url{https://arxiv.org/abs/1810.01943}.

\bibitem[Bhatt et~al.(2021)Bhatt, Antor{\'a}n, Zhang, Liao, Sattigeri,
  Fogliato, Melan{\c{c}}on, Krishnan, Stanley, Tickoo,
  et~al.]{bhatt2021uncertainty}
Umang Bhatt, Javier Antor{\'a}n, Yunfeng Zhang, Q~Vera Liao, Prasanna
  Sattigeri, Riccardo Fogliato, Gabrielle Melan{\c{c}}on, Ranganath Krishnan,
  Jason Stanley, Omesh Tickoo, et~al.
\newblock Uncertainty as a form of transparency: Measuring, communicating, and
  using uncertainty.
\newblock In \emph{Proceedings of the 2021 AAAI/ACM Conference on AI, Ethics,
  and Society}, pp.\  401--413, 2021.

\bibitem[Bird et~al.(2020)Bird, Dud{\'i}k, Edgar, Horn, Lutz, Milan, Sameki,
  Wallach, and Walker]{bird2020fairlearn}
Sarah Bird, Miro Dud{\'i}k, Richard Edgar, Brandon Horn, Roman Lutz, Vanessa
  Milan, Mehrnoosh Sameki, Hanna Wallach, and Kathleen Walker.
\newblock Fairlearn: A toolkit for assessing and improving fairness in {AI}.
\newblock Technical Report MSR-TR-2020-32, Microsoft, May 2020.
\newblock URL
  \url{https://www.microsoft.com/en-us/research/publication/fairlearn-a-toolkit-for-assessing-and-improving-fairness-in-ai/}.

\bibitem[Black et~al.(2022{\natexlab{a}})Black, Leino, and
  Fredrikson]{black2021selective}
Emily Black, Klas Leino, and Matt Fredrikson.
\newblock Selective ensembles for consistent predictions.
\newblock In \emph{\ICLR{2022}}, 2022{\natexlab{a}}.

\bibitem[Black et~al.(2022{\natexlab{b}})Black, Raghavan, and
  Barocas]{black2022model}
Emily Black, Manish Raghavan, and Solon Barocas.
\newblock Model multiplicity: Opportunities, concerns, and solutions.
\newblock In \emph{2022 ACM Conference on Fairness, Accountability, and
  Transparency}, pp.\  850--863, 2022{\natexlab{b}}.

\bibitem[Caton \& Haas(2024)Caton and Haas]{caton2020fairness}
Simon Caton and Christian Haas.
\newblock Fairness in machine learning: A survey.
\newblock \emph{ACM Comput. Surv.}, 56\penalty0 (7), April 2024.
\newblock ISSN 0360-0300.
\newblock \doi{10.1145/3616865}.
\newblock URL \url{https://doi.org/10.1145/3616865}.

\bibitem[Chen et~al.(2018)Chen, Johansson, and Sontag]{chen2018my}
Irene Chen, Fredrik~D Johansson, and David Sontag.
\newblock Why is my classifier discriminatory?
\newblock \emph{Advances in neural information processing systems}, 31, 2018.

\bibitem[Chen \& Guestrin(2016)Chen and Guestrin]{chen2016xgboost}
Tianqi Chen and Carlos Guestrin.
\newblock Xgboost: A scalable tree boosting system.
\newblock In \emph{Proceedings of the 22nd acm sigkdd international conference
  on knowledge discovery and data mining}, pp.\  785--794, 2016.

\bibitem[Collier et~al.(2020)Collier, Mustafa, Kokiopoulou, Jenatton, and
  Berent]{collier2020simple}
Mark Collier, Basil Mustafa, Efi Kokiopoulou, Rodolphe Jenatton, and Jesse
  Berent.
\newblock A simple probabilistic method for deep classification under
  input-dependent label noise.
\newblock \emph{arXiv preprint arXiv:2003.06778}, 2020.

\bibitem[Cooper et~al.(2024)Cooper, Lee, Choksi, Barocas, De~Sa, Grimmelmann,
  Kleinberg, Sen, and Zhang]{cooper2024arbitrariness}
A~Feder Cooper, Katherine Lee, Madiha~Zahrah Choksi, Solon Barocas, Christopher
  De~Sa, James Grimmelmann, Jon Kleinberg, Siddhartha Sen, and Baobao Zhang.
\newblock Arbitrariness and social prediction: The confounding role of variance
  in fair classification.
\newblock In \emph{Proceedings of the AAAI Conference on Artificial
  Intelligence}, volume~38, pp.\  22004--22012, 2024.

\bibitem[Corbett-Davies et~al.(2023)Corbett-Davies, Gaebler, Nilforoshan,
  Shroff, and Goel]{corbett2023measure}
Sam Corbett-Davies, Johann~D Gaebler, Hamed Nilforoshan, Ravi Shroff, and
  Sharad Goel.
\newblock The measure and mismeasure of fairness.
\newblock \emph{The Journal of Machine Learning Research}, 24\penalty0
  (1):\penalty0 14730--14846, 2023.

\bibitem[Ding et~al.(2021)Ding, Hardt, Miller, and Schmidt]{ding2021retiring}
Frances Ding, Moritz Hardt, John Miller, and Ludwig Schmidt.
\newblock Retiring adult: New datasets for fair machine learning.
\newblock \emph{Advances in neural information processing systems},
  34:\penalty0 6478--6490, 2021.

\bibitem[Fabris et~al.(2022)Fabris, Messina, Silvello, and
  Susto]{fabris2022algorithmic}
Alessandro Fabris, Stefano Messina, Gianmaria Silvello, and Gian~Antonio Susto.
\newblock Algorithmic fairness datasets: the story so far.
\newblock \emph{Data Mining and Knowledge Discovery}, 36\penalty0 (6):\penalty0
  2074--2152, 2022.

\bibitem[Fisher(1925)]{fisher1925theory}
Ronald~Aylmer Fisher.
\newblock Theory of statistical estimation.
\newblock In \emph{Mathematical proceedings of the Cambridge philosophical
  society}, volume~22, pp.\  700--725. Cambridge University Press, 1925.

\bibitem[Gorishniy et~al.(2021)Gorishniy, Rubachev, Khrulkov, and
  Babenko]{gorishniy2021revisiting}
Yury Gorishniy, Ivan Rubachev, Valentin Khrulkov, and Artem Babenko.
\newblock Revisiting deep learning models for tabular data.
\newblock \emph{Advances in Neural Information Processing Systems},
  34:\penalty0 18932--18943, 2021.

\bibitem[Grinsztajn et~al.(2022)Grinsztajn, Oyallon, and
  Varoquaux]{grinsztajn2022tree}
L{\'e}o Grinsztajn, Edouard Oyallon, and Ga{\"e}l Varoquaux.
\newblock Why do tree-based models still outperform deep learning on typical
  tabular data?
\newblock \emph{Advances in neural information processing systems},
  35:\penalty0 507--520, 2022.

\bibitem[Guo et~al.(2017)Guo, Pleiss, Sun, and Weinberger]{guo2017calibration}
Chuan Guo, Geoff Pleiss, Yu~Sun, and Kilian~Q Weinberger.
\newblock On calibration of modern neural networks.
\newblock In \emph{International conference on machine learning}, pp.\
  1321--1330. PMLR, 2017.

\bibitem[Han et~al.(2022)Han, Liang, Yang, Liu, Li, Bian, Zhao, Wu, Zhang, and
  Yao]{han2022umix}
Zongbo Han, Zhipeng Liang, Fan Yang, Liu Liu, Lanqing Li, Yatao Bian, Peilin
  Zhao, Bingzhe Wu, Changqing Zhang, and Jianhua Yao.
\newblock Umix: Improving importance weighting for subpopulation shift via
  uncertainty-aware mixup.
\newblock \emph{Advances in Neural Information Processing Systems},
  35:\penalty0 37704--37718, 2022.

\bibitem[Hardt et~al.(2016)Hardt, Price, and Srebro]{hardt2016equality}
Moritz Hardt, Eric Price, and Nati Srebro.
\newblock Equality of opportunity in supervised learning.
\newblock \emph{Advances in neural information processing systems}, 29, 2016.

\bibitem[Hayes \& Cai(2007)Hayes and Cai]{hayes2007using}
Andrew~F Hayes and Li~Cai.
\newblock Using heteroskedasticity-consistent standard error estimators in ols
  regression: An introduction and software implementation.
\newblock \emph{Behavior research methods}, 39:\penalty0 709--722, 2007.

\bibitem[Hendrickx et~al.(2024)Hendrickx, Perini, Van~der Plas, Meert, and
  Davis]{hendrickx2024machine}
Kilian Hendrickx, Lorenzo Perini, Dries Van~der Plas, Wannes Meert, and Jesse
  Davis.
\newblock Machine learning with a reject option: A survey.
\newblock \emph{Machine Learning}, 113\penalty0 (5):\penalty0 3073--3110, 2024.

\bibitem[Hill(2011)]{hill2011bayesian}
Jennifer~L Hill.
\newblock Bayesian nonparametric modeling for causal inference.
\newblock \emph{Journal of Computational and Graphical Statistics}, 20\penalty0
  (1):\penalty0 217--240, 2011.

\bibitem[Hofmann(2000)]{hofmann2000institut}
H~Hofmann.
\newblock Institut f{\"u}r statstik und okonometrie universit{\"a}t, 2000.

\bibitem[H{\"u}llermeier \& Waegeman(2021)H{\"u}llermeier and
  Waegeman]{hullermeier2021aleatoric}
Eyke H{\"u}llermeier and Willem Waegeman.
\newblock Aleatoric and epistemic uncertainty in machine learning: An
  introduction to concepts and methods.
\newblock \emph{Machine learning}, 110\penalty0 (3):\penalty0 457--506, 2021.

\bibitem[Ji et~al.(2020)Ji, Smyth, and Steyvers]{ji2020can}
Disi Ji, Padhraic Smyth, and Mark Steyvers.
\newblock Can i trust my fairness metric? assessing fairness with unlabeled
  data and bayesian inference.
\newblock \emph{Advances in Neural Information Processing Systems},
  33:\penalty0 18600--18612, 2020.

\bibitem[Kohavi et~al.(1996)]{kohavi1996scaling}
Ron Kohavi et~al.
\newblock Scaling up the accuracy of naive-bayes classifiers: A decision-tree
  hybrid.
\newblock In \emph{Kdd}, volume~96, pp.\  202--207, 1996.

\bibitem[Kuzucu et~al.(2023)Kuzucu, Cheong, Gunes, and
  Kalkan]{kuzucu2023uncertainty}
Selim Kuzucu, Jiaee Cheong, Hatice Gunes, and Sinan Kalkan.
\newblock Uncertainty-based fairness measures.
\newblock \emph{arXiv preprint arXiv:2312.11299}, 2023.

\bibitem[Lakshminarayanan et~al.(2017)Lakshminarayanan, Pritzel, and
  Blundell]{lakshminarayanan2017simple}
Balaji Lakshminarayanan, Alexander Pritzel, and Charles Blundell.
\newblock Simple and scalable predictive uncertainty estimation using deep
  ensembles.
\newblock \emph{Advances in neural information processing systems}, 30, 2017.

\bibitem[Lantz(2019)]{lantz2019machine}
Brett Lantz.
\newblock \emph{Machine learning with R: expert techniques for predictive
  modeling}.
\newblock Packt publishing ltd, 2019.

\bibitem[Lee(2018)]{lee2018detecting}
Nicol~Turner Lee.
\newblock Detecting racial bias in algorithms and machine learning.
\newblock \emph{Journal of Information, Communication and Ethics in Society},
  2018.

\bibitem[Liu et~al.(2022)Liu, Ding, Yu, Liu, Kong, and
  Jiang]{liu2022conformalized}
Meichen Liu, Lei Ding, Dengdeng Yu, Wulong Liu, Linglong Kong, and Bei Jiang.
\newblock Conformalized fairness via quantile regression.
\newblock \emph{Advances in Neural Information Processing Systems},
  35:\penalty0 11561--11572, 2022.

\bibitem[Long et~al.(2024)Long, Hsu, Alghamdi, and Calmon]{long2024individual}
Carol Long, Hsiang Hsu, Wael Alghamdi, and Flavio Calmon.
\newblock Individual arbitrariness and group fairness.
\newblock \emph{Advances in Neural Information Processing Systems}, 36, 2024.

\bibitem[MacKinnon(2012)]{mackinnon2012thirty}
James~G MacKinnon.
\newblock Thirty years of heteroskedasticity-robust inference.
\newblock In \emph{Recent advances and future directions in causality,
  prediction, and specification analysis: Essays in honor of Halbert L. White
  Jr}, pp.\  437--461. Springer, 2012.

\bibitem[Madras et~al.(2018)Madras, Pitassi, and Zemel]{madras2018predict}
David Madras, Toni Pitassi, and Richard Zemel.
\newblock Predict responsibly: improving fairness and accuracy by learning to
  defer.
\newblock \emph{Advances in neural information processing systems}, 31, 2018.

\bibitem[Malinin \& Gales(2018)Malinin and Gales]{malinin2018predictive}
Andrey Malinin and Mark Gales.
\newblock Predictive uncertainty estimation via prior networks.
\newblock \emph{Advances in neural information processing systems}, 31, 2018.

\bibitem[McElfresh et~al.(2024)McElfresh, Khandagale, Valverde, Prasad~C,
  Ramakrishnan, Goldblum, and White]{mcelfresh2024neural}
Duncan McElfresh, Sujay Khandagale, Jonathan Valverde, Vishak Prasad~C, Ganesh
  Ramakrishnan, Micah Goldblum, and Colin White.
\newblock When do neural nets outperform boosted trees on tabular data?
\newblock \emph{Advances in Neural Information Processing Systems}, 36, 2024.

\bibitem[McNair(2018)]{mcnair2018preventing}
Douglas~S McNair.
\newblock Preventing disparities: Bayesian and frequentist methods for
  assessing fairness in machinelearning decision-support models.
\newblock \emph{New Insights into Bayesian Inference}, 71, 2018.

\bibitem[Mitchell et~al.(2021)Mitchell, Potash, Barocas, D'Amour, and
  Lum]{mitchell2021algorithmic}
Shira Mitchell, Eric Potash, Solon Barocas, Alexander D'Amour, and Kristian
  Lum.
\newblock Algorithmic fairness: Choices, assumptions, and definitions.
\newblock \emph{Annual review of statistics and its application}, 8:\penalty0
  141--163, 2021.

\bibitem[Moro et~al.(2014)Moro, Cortez, and Rita]{moro2014data}
S{\'e}rgio Moro, Paulo Cortez, and Paulo Rita.
\newblock A data-driven approach to predict the success of bank telemarketing.
\newblock \emph{Decision Support Systems}, 62:\penalty0 22--31, 2014.

\bibitem[Redmond \& Baveja(2002)Redmond and Baveja]{redmond2002data}
Michael Redmond and Alok Baveja.
\newblock A data-driven software tool for enabling cooperative information
  sharing among police departments.
\newblock \emph{European Journal of Operational Research}, 141\penalty0
  (3):\penalty0 660--678, 2002.

\bibitem[Rigobon(2003)]{rigobon2003identification}
Roberto Rigobon.
\newblock Identification through heteroskedasticity.
\newblock \emph{Review of Economics and Statistics}, 85\penalty0 (4):\penalty0
  777--792, 2003.

\bibitem[Sander(2004)]{sander2004systemic}
Richard~H Sander.
\newblock A systemic analysis of affirmative action in american law schools.
\newblock \emph{Stan. L. Rev.}, 57:\penalty0 367, 2004.

\bibitem[Seitzer et~al.(2022)Seitzer, Tavakoli, Antic, and
  Martius]{seitzer2022pitfalls}
Maximilian Seitzer, Arash Tavakoli, Dimitrije Antic, and Georg Martius.
\newblock On the pitfalls of heteroscedastic uncertainty estimation with
  probabilistic neural networks.
\newblock \emph{arXiv preprint arXiv:2203.09168}, 2022.

\bibitem[Shah et~al.(2022)Shah, Bu, Lee, Das, Panda, Sattigeri, and
  Wornell]{shah2022selective}
Abhin Shah, Yuheng Bu, Joshua~K Lee, Subhro Das, Rameswar Panda, Prasanna
  Sattigeri, and Gregory~W Wornell.
\newblock Selective regression under fairness criteria.
\newblock In \emph{International Conference on Machine Learning}, pp.\
  19598--19615. PMLR, 2022.

\bibitem[Stirn et~al.(2023)Stirn, Wessels, Schertzer, Pereira, Sanjana, and
  Knowles]{stirn2023faithful}
Andrew Stirn, Harm Wessels, Megan Schertzer, Laura Pereira, Neville Sanjana,
  and David Knowles.
\newblock Faithful heteroscedastic regression with neural networks.
\newblock In \emph{International Conference on Artificial Intelligence and
  Statistics}, pp.\  5593--5613. PMLR, 2023.

\bibitem[Tahir et~al.(2023)Tahir, Cheng, and Liu]{tahir2023fairness}
Anique Tahir, Lu~Cheng, and Huan Liu.
\newblock Fairness through aleatoric uncertainty.
\newblock In \emph{Proceedings of the 32nd ACM International Conference on
  Information and Knowledge Management}, pp.\  2372--2381, 2023.

\bibitem[Thomas \& Pont{\'o}n-N{\'u}{\~n}ez(2022)Thomas and
  Pont{\'o}n-N{\'u}{\~n}ez]{thomas2022automating}
Christopher Thomas and Antonio Pont{\'o}n-N{\'u}{\~n}ez.
\newblock Automating judicial discretion: How algorithmic risk assessments in
  pretrial adjudications violate equal protection rights on the basis of race.
\newblock \emph{Law \& Ineq.}, 40:\penalty0 371, 2022.

\bibitem[Wang et~al.(2024{\natexlab{a}})Wang, Cheng, Guo, Liu, and
  Yu]{wang2024equal}
Fangxin Wang, Lu~Cheng, Ruocheng Guo, Kay Liu, and Philip~S Yu.
\newblock Equal opportunity of coverage in fair regression.
\newblock \emph{Advances in Neural Information Processing Systems}, 36,
  2024{\natexlab{a}}.

\bibitem[Wang et~al.(2024{\natexlab{b}})Wang, He, Gao, and
  Calmon]{wang2024aleatoric}
Hao Wang, Luxi He, Rui Gao, and Flavio Calmon.
\newblock Aleatoric and epistemic discrimination: Fundamental limits of
  fairness interventions.
\newblock \emph{Advances in Neural Information Processing Systems}, 36,
  2024{\natexlab{b}}.

\bibitem[Wang \& Wang(2024)Wang and Wang]{wang2024variational}
Ziyan Wang and Hao Wang.
\newblock Variational imbalanced regression: Fair uncertainty quantification
  via probabilistic smoothing.
\newblock \emph{Advances in Neural Information Processing Systems}, 36, 2024.

\bibitem[White(1980)]{white1980heteroskedasticity}
Halbert White.
\newblock A heteroskedasticity-consistent covariance matrix estimator and a
  direct test for heteroskedasticity.
\newblock \emph{Econometrica: journal of the Econometric Society}, pp.\
  817--838, 1980.

\bibitem[Zafar et~al.(2017)Zafar, Valera, Gomez~Rodriguez, and
  Gummadi]{zafar2017fairness}
Muhammad~Bilal Zafar, Isabel Valera, Manuel Gomez~Rodriguez, and Krishna~P
  Gummadi.
\newblock Fairness beyond disparate treatment \& disparate impact: Learning
  classification without disparate mistreatment.
\newblock In \emph{Proceedings of the 26th international conference on world
  wide web}, pp.\  1171--1180, 2017.

\bibitem[Zhang \& Long(2021)Zhang and Long]{zhang2021assessing}
Yiliang Zhang and Qi~Long.
\newblock Assessing fairness in the presence of missing data.
\newblock \emph{Advances in neural information processing systems},
  34:\penalty0 16007--16019, 2021.

\end{thebibliography}
\bibliographystyle{references_style}

\newpage
\appendix
\section{Datasets Included in \FU}

\begin{table}[h!]
\centering
\caption{We note that even though ACS Folktables is the improved version of Adult, we include Adult for parity with prior work~\citep{ding2021retiring,cooper2024arbitrariness}; we include COMPAS and Communities \& Crimes for similar parity reasons, although highlight significant concerns with automating criminal justice~\citep{fabris2022algorithmic, thomas2022automating}.}
\label{table:datasets}
\resizebox{\linewidth}{!}{
\begin{tabular}{llllll}
\toprule
\midrule
\textbf{Datasets (binary)} & \textbf{Size} & \textbf{\# Feat.} & \textbf{Goal is to predict...} & \textbf{Protected Att.} \\ 
\midrule
ACS \cite{ding2021retiring} & 16249 & 16 & whether an individual is employed & race \\ 
Adult \cite{kohavi1996scaling} & 45222 & 102 & whether individual income exceeds a certain level & gender \\ 
Bank Marketing \cite{moro2014data} & 30488 & 57 & whether clients will subscribe to a product & age \\ 
COMPAS \cite{angwin2022machine} & 6167 & 406 & whether a defendant will re-offend & gender \\ 
German Credit \cite{hofmann2000institut} & 1000 & 57 & whether an individual has `good' or `bad' credit & age \\ 
\midrule
\midrule
\textbf{Datasets (regression)} &  &  &  &  \\ 
\midrule
Law School GPA \cite{sander2004systemic} & 22342 & 4 & students' GPA in law school & race \\ 
Communities \& Crimes \cite{redmond2002data} & 1994 & 100 & \# of per-capita violent crimes in a community & race \\ 
Insurance \cite{lantz2019machine} & 1338 & 8 & individual medical costs billed by insurance & gender \\ 
IHDP \cite{hill2011bayesian} & 747 & 26 & the cognitive test scores of infants & gender \\ 
Twins \cite{almond2005costs} & 68995 & 19 & the number of prenatal visits & race \\ 
\midrule
\bottomrule
\end{tabular}}
\vspace{-1em}
\end{table}

\clearpage
\section{Additional Results on Consistency and Calibration}\label{appendix:consistency}

\paragraph{Why to Use \textit{Binomial NLL} for Binary Classification Uncertainty} Consider a model $\tilde{f}: \mathcal{X} \times \mathcal{A} \to \mathcal{Y} \times \mathbb{R}$ that outputs binary predictions $\tilde{y}$ and uncertainty estimates $\tilde{\sigma}$.
We can evaluate the NLL of the estimates by converting them to probabilities. 
We can solve for probabilities that produce the estimates $\tilde{\sigma}$ with the quadratic formula:\footnote{The catch is that two probabilities can map to the same standard deviation.
Since the model $\tilde{f}$ also makes predictions $\tilde{y}$, we will choose the probability $\tilde{p}$ closer to the prediction.}
\begin{align*}
\tilde{p}^- = \frac{1 - \sqrt{1 - 4\tilde{\sigma}^2}}{2}
&&
\tilde{p}^+ = \frac{1 + \sqrt{1 - 4\tilde{\sigma}^2}}{2}.
\end{align*}
Implicitly, \textit{Binomial NLL} argues that if we had some $(\tilde{\sigma}’)^2$ that was more accurate than the estimated uncertainty $\tilde{\sigma}$,
then we could get \textit{more accurate probability predictions} $\tilde{p}’$ \textit{from} $(\tilde{\sigma}')^2$. Thus, we should simply do our best to estimate $\tilde{p}$ (with NLL as the natural objective), which then induces a standard deviation $\tilde{\sigma}$. This motivates \textit{Binomial NLL} and likely explains its strong performance.

\begin{figure}[h!]
    \centering
    \includegraphics[width=\linewidth]{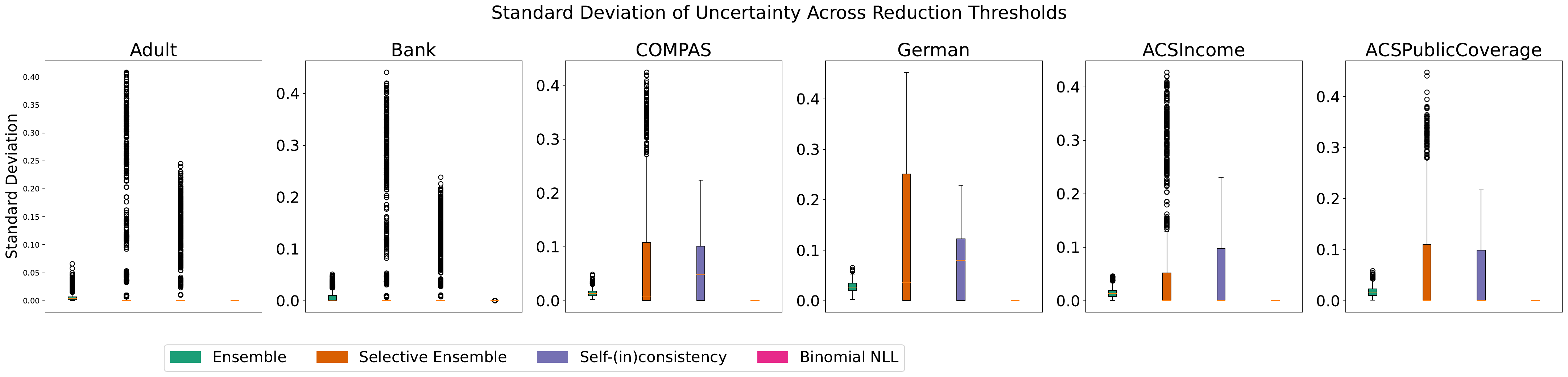}
    \caption{Boxplot displaying the mean, variance, and outliers of individual uncertainty predictions for the binary classification setting, with varying the \textit{reduction\_threshold} hyperparameter $\gamma$ (bottom plot) for the XGBoost model. The Binomial NLL method demonstrates significantly greater consistency across varying $\gamma$ values.}
    \label{fig:binary_consistency_gamma}
\end{figure}

\begin{figure}[h!]
    \centering
    \includegraphics[width=\linewidth]{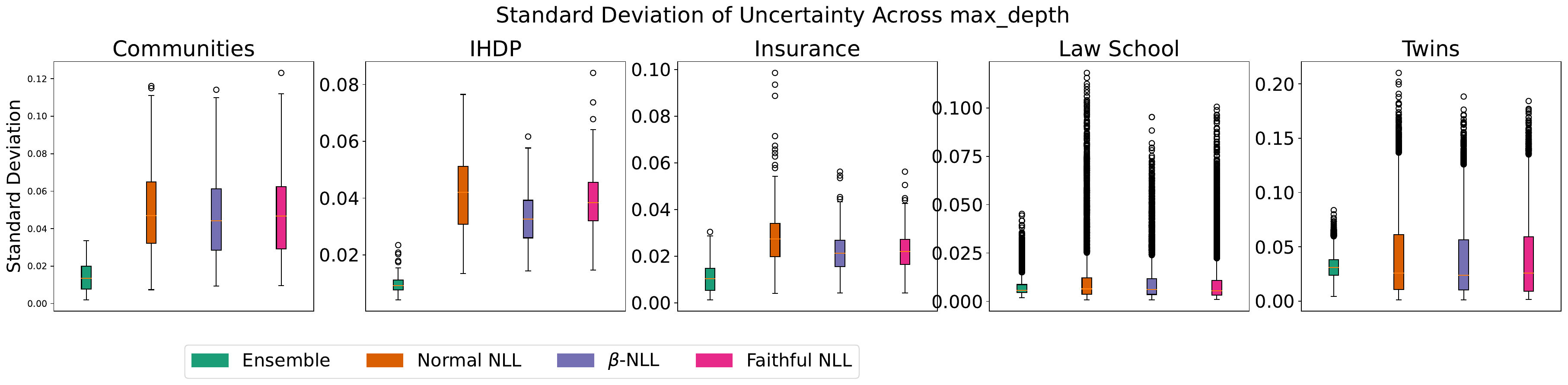}
    \includegraphics[width=\linewidth]{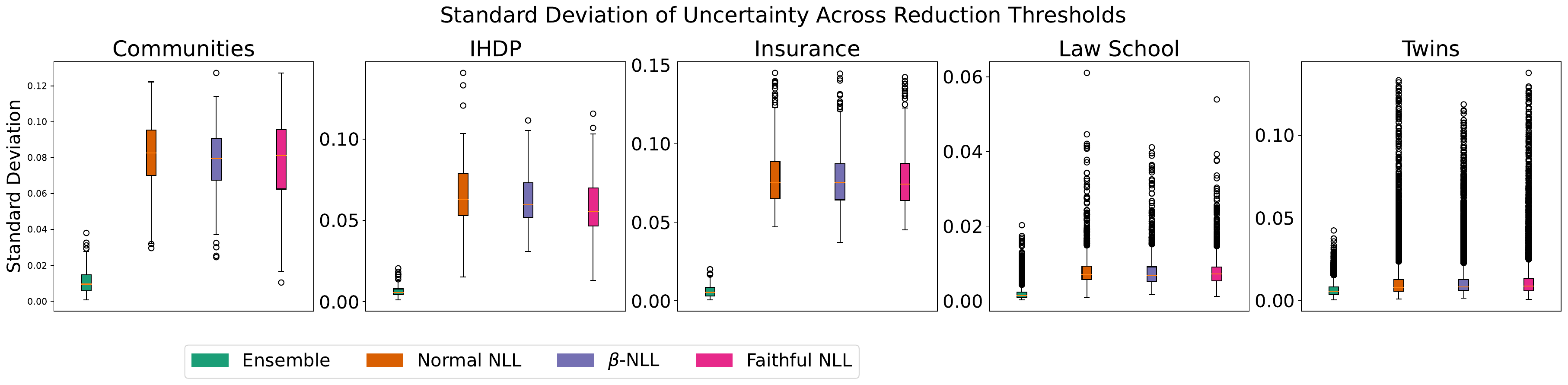}
    \caption{Boxplots showing the variance of individual uncertainty predictions for the \textbf{regression} setting with varying max$\_$depth (top plot) and reduction threshold $\gamma$ (bottom plot) parameters. The Ensemble method is the most consistent across the max$\_$depth values, while all other methods exhibit similar levels of consistency.}
    \label{fig:regression_consistency}
\end{figure}

\begin{figure}[h]
    \centering
    \includegraphics[width=\linewidth]{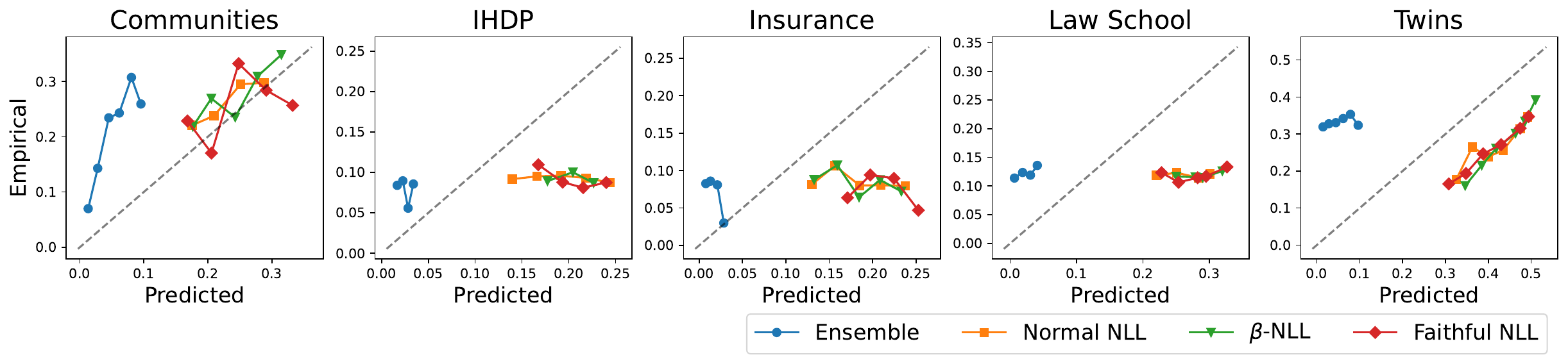}
    \caption{The \emph{calibration} of various algorithms. We compute \emph{calibration} by making 100 bins on the uncertainty measure: we plot the empirical standard standard deviation in the bin against the predicted uncertainty.}
    \label{fig:regression_calibration}
\end{figure}

\begin{figure}[h!]
    \centering
    \includegraphics[width=.8\linewidth]{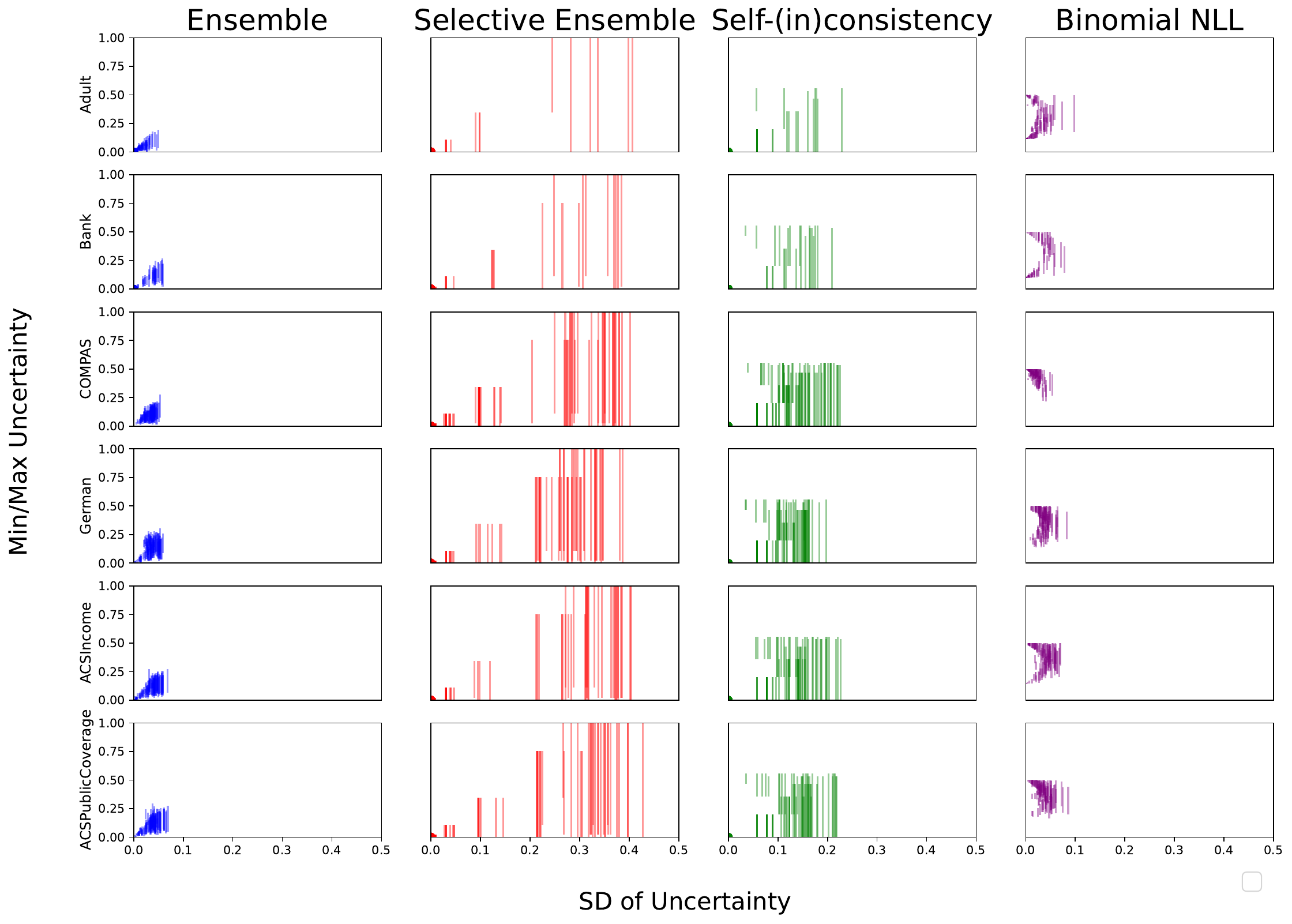}
    \caption{Plots of uncertainty for the binary classification problem across datasets (rows) and methods (columns), where we vary the hyperparameter \textit{max-depth}. Each individual is plotted as a line, where the line's higher y-coordinate is the \textit{maximum} uncertainty estimate over all hyperparameter settings for that, and the lower y-coordinate is the individual's \textit{minimum} uncertainty estimate. Each individual is also placed along the $x$-axis at the standard deviation of their uncertainty estimates over the hyper-parameter settings. The \textit{Ensemble} method produces \textit{consistent} uncertainty estimates: the standard deviation of these estimates is less than 0.1 for all datasets and the uncertainty range starts close to 0 and increases with the standard deviation of the uncertainty (this is desirable). In contrast, the \textit{Selective Ensemble} method produces \textit{inconsistent} uncertainty estimates: the standard deviation of the uncertainty predictions is large and the range of the uncertainty predictions is large (especially for large standard deviations).}
    \label{fig:consistency_range}
\end{figure}

\paragraph{Additional experiments} In Figure~\ref{fig:binary_consistency_gamma}, we present boxplots of individual uncertainty predictions for binary classification, varying the reduction threshold $\gamma$ for the XGBoost model; the \textit{Binomial NLL} method demonstrates greater consistency across different $\gamma$ values. Figure~\ref{fig:regression_consistency} shows similar boxplots in the regression setting, varying \textit{max\_depth} (top plot) and reduction threshold $\gamma$ (bottom plot), indicating that the \textit{Ensemble} method is the most consistent across \textit{max\_depth} values, while other methods exhibit similar consistency levels. Figure~\ref{fig:regression_calibration} illustrates the calibration of various algorithms in regression by plotting the empirical standard deviation against the predicted uncertainty using 100 bins. Finally, Figure~\ref{fig:consistency_range} displays uncertainty plots for binary classification across datasets and methods, varying \textit{max\_depth}, demonstrating that the \textit{Ensemble} method produces consistent uncertainty estimates, whereas the \textit{Selective Ensemble} method shows large variance and inconsistency in uncertainty estimates.

\clearpage
\section{Binary Abstention Results}\label{appendix:abstain}

\begin{figure}[h]
    \centering
    \includegraphics[width=\linewidth]{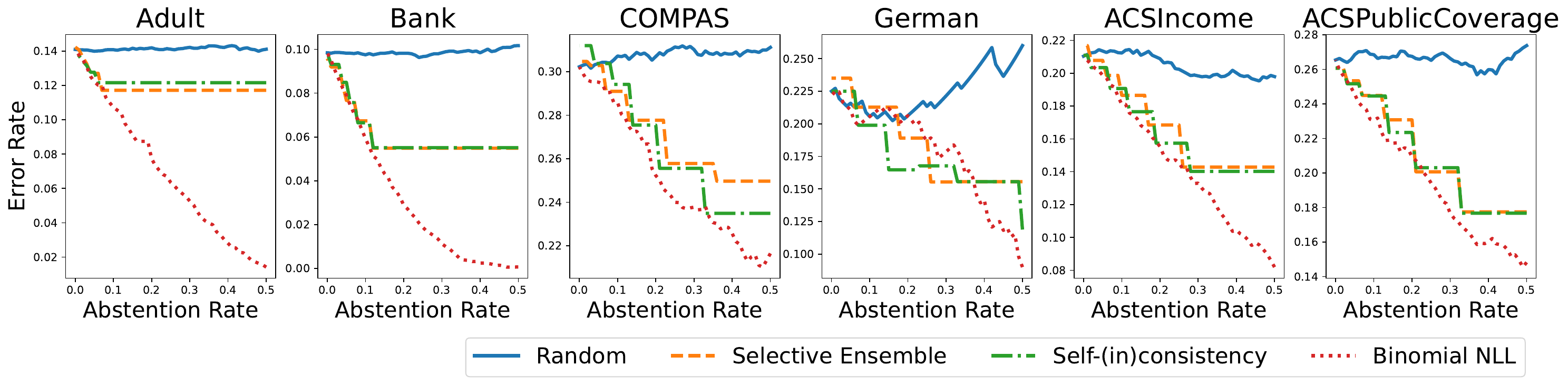}
    \caption{Abstention error rate.}
    \label{fig:abstention_er}
\end{figure}

\begin{figure}[h]
    \centering
    \includegraphics[width=\linewidth]{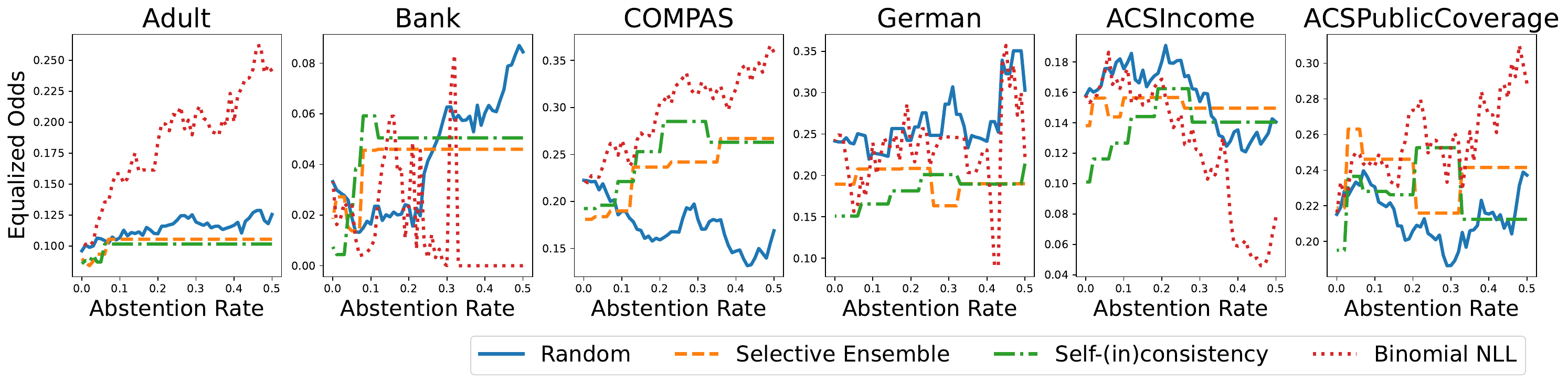}
    \caption{Abstention equalized odds.}
    \label{fig:abstention_eo}
\end{figure}

\begin{figure}[h]
    \centering
    \includegraphics[width=\linewidth]{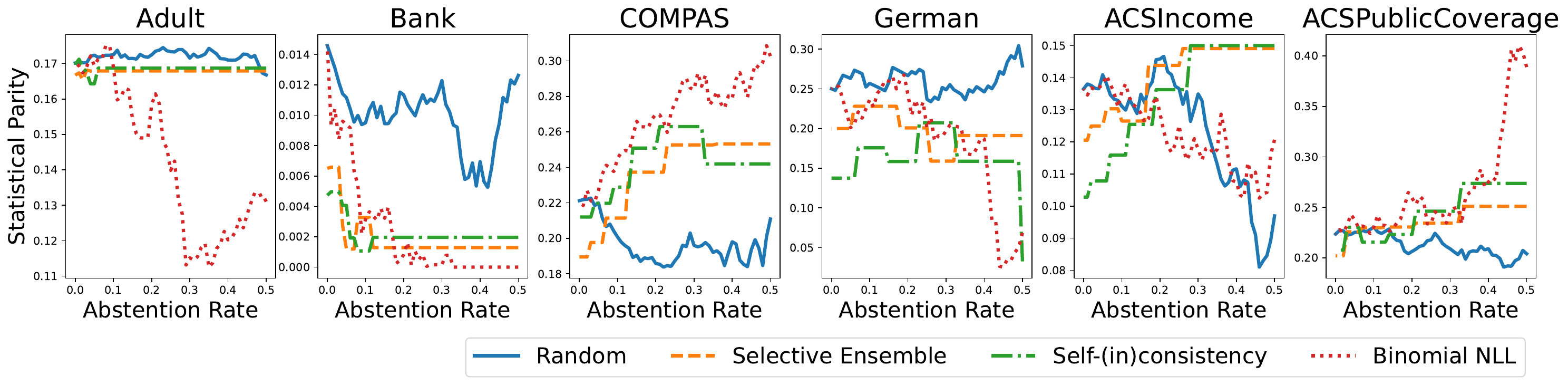}
    \caption{Abstention statistical parity.}
    \label{fig:abstention_sp}
\end{figure}

\begin{figure}[h!]
    \centering
    \includegraphics[width=0.85\linewidth]{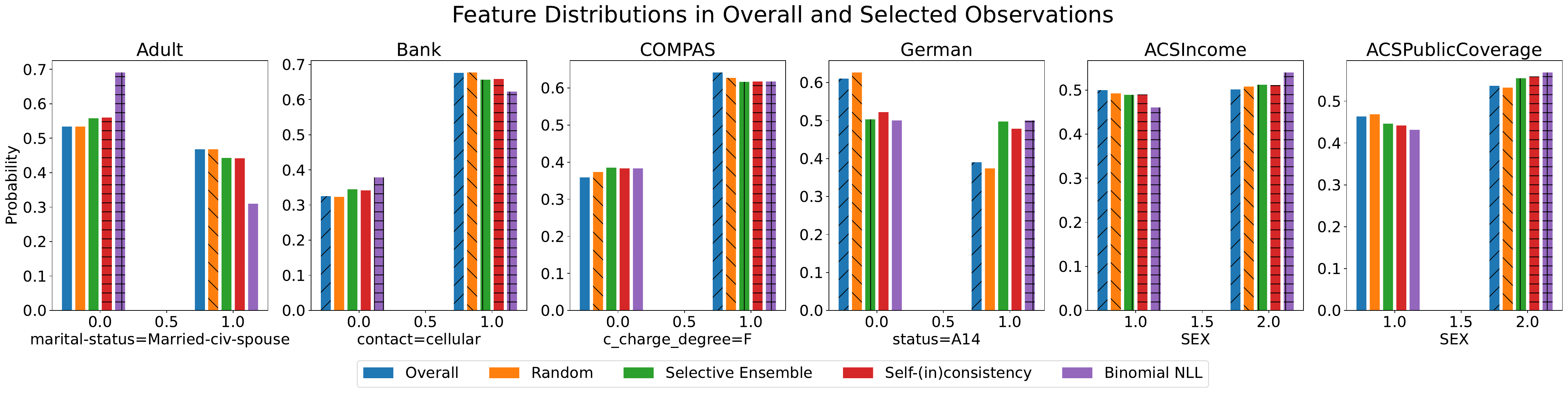}
    \caption{The variable with the largest difference (as measured by \textit{Wasserstein} distance) between the distribution on the overall population and the set selected through the abstaining process. Except for the \textit{Binomial NLL} distribution on the \textit{Adult} dataset, the differences tend to be relatively small.}
    \label{fig:wasserstein}
\end{figure}

\FU explores how abstaining from binary predictions based on uncertainty affects error rate, equalized odds, and statistical parity. While it clearly reduces error rate (Figure~\ref{fig:abstention_er}), abstaining has an unreliable effect on equalized odds (Figure~\ref{fig:abstention_eo}) and statistical parity (Figure~\ref{fig:abstention_sp}), similar to the behavior of the \textit{Random} baseline. Figure~\ref{fig:wasserstein} shows that the differences in variable distributions between the overall population and the set selected through the abstaining process are relatively small, except for the \textit{Binomial NLL} distribution on the \textit{Adult} dataset.

\clearpage

\section{Binary Fairness Results}\label{appendix:binary_fair}

\FU fairness benchmark on all five binary datasets.
Abstaining reduces the error rate, and often equalized odds as a result, but has no improvement on statistical parity. Only \textit{Predictive Parity} (the difference in the ratio of true positives to all positive labels assigned) was relatively low for abstaining methods, and is likely explained by none of the fairness methods explicitly optimizing this parameter.

\begin{table}[h]
    \centering
    \resizebox{\textwidth}{!}{\begin{tabular} {lcccccccc}
\toprule
\textbf{Approach} & \textbf{Error Rate} & \textbf{Statistical Parity} & \textbf{Equalized Odds} & \textbf{Equal Opportunity} & \textbf{Disparate Impact} & \textbf{Predictive Parity} & \textbf{False Positive Rate} & \textbf{Included \%} \\ \midrule
Baseline & 0.22 $\pm$ 0.009 & 0.181 $\pm$ 0.023 & 0.165 $\pm$ 0.049 & 0.165 $\pm$ 0.049 & 1.65 $\pm$ 0.136 & 0.057 $\pm$ 0.034 & 0.083 $\pm$ 0.024 & \cellcolor{gold!30}100.0 $\pm$ 0.0 \\ 
Threshold Optimizer SP & 0.233 $\pm$ 0.009 & \cellcolor{gold!30}0.029 $\pm$ 0.021 & 0.135 $\pm$ 0.046 & \cellcolor{gold!30}0.063 $\pm$ 0.033 & \cellcolor{silver!30}0.979 $\pm$ 0.07 & 0.241 $\pm$ 0.043 & 0.134 $\pm$ 0.046 & \cellcolor{gold!30}100.0 $\pm$ 0.0 \\ 
Threshold Optimizer EO & 0.226 $\pm$ 0.01 & 0.101 $\pm$ 0.03 & \cellcolor{gold!30}0.08 $\pm$ 0.049 & \cellcolor{bronze!30}0.079 $\pm$ 0.049 & 1.3 $\pm$ 0.113 & 0.119 $\pm$ 0.039 & \cellcolor{gold!30}0.024 $\pm$ 0.021 & \cellcolor{gold!30}100.0 $\pm$ 0.0 \\ 
Exponentiated Gradient SP & 0.229 $\pm$ 0.011 & \cellcolor{silver!30}0.041 $\pm$ 0.022 & \cellcolor{bronze!30}0.117 $\pm$ 0.038 & \cellcolor{silver!30}0.065 $\pm$ 0.04 & \cellcolor{gold!30}0.946 $\pm$ 0.08 & 0.203 $\pm$ 0.033 & 0.115 $\pm$ 0.038 & \cellcolor{gold!30}100.0 $\pm$ 0.0 \\ 
Exponentiated Gradient EO & 0.218 $\pm$ 0.008 & 0.101 $\pm$ 0.041 & \cellcolor{silver!30}0.092 $\pm$ 0.049 & 0.09 $\pm$ 0.052 & 1.3 $\pm$ 0.15 & 0.119 $\pm$ 0.045 & \cellcolor{silver!30}0.031 $\pm$ 0.017 & \cellcolor{gold!30}100.0 $\pm$ 0.0 \\ 
Grid Search SP & 0.23 $\pm$ 0.012 & \cellcolor{bronze!30}0.083 $\pm$ 0.045 & 0.142 $\pm$ 0.064 & 0.108 $\pm$ 0.067 & \cellcolor{bronze!30}1.04 $\pm$ 0.259 & 0.183 $\pm$ 0.048 & 0.106 $\pm$ 0.067 & \cellcolor{gold!30}100.0 $\pm$ 0.0 \\ 
Grid Search EO & 0.227 $\pm$ 0.008 & 0.148 $\pm$ 0.034 & 0.132 $\pm$ 0.07 & 0.125 $\pm$ 0.074 & 1.48 $\pm$ 0.152 & 0.087 $\pm$ 0.048 & \cellcolor{bronze!30}0.056 $\pm$ 0.031 & \cellcolor{gold!30}100.0 $\pm$ 0.0 \\ 
\midrule
Random & 0.221 $\pm$ 0.009 & 0.174 $\pm$ 0.028 & 0.154 $\pm$ 0.046 & 0.15 $\pm$ 0.052 & 1.61 $\pm$ 0.155 & 0.044 $\pm$ 0.034 & 0.085 $\pm$ 0.024 & 88.3 $\pm$ 6.96 \\ 
Ensemble & \cellcolor{bronze!30}0.2 $\pm$ 0.019 & 0.193 $\pm$ 0.032 & 0.158 $\pm$ 0.063 & 0.155 $\pm$ 0.064 & 1.72 $\pm$ 0.18 & \cellcolor{gold!30}0.032 $\pm$ 0.038 & 0.093 $\pm$ 0.031 & 89.0 $\pm$ 8.04 \\ 
Selective Ensemble & 0.201 $\pm$ 0.024 & 0.183 $\pm$ 0.03 & 0.154 $\pm$ 0.049 & 0.15 $\pm$ 0.059 & 1.69 $\pm$ 0.204 & 0.039 $\pm$ 0.037 & 0.082 $\pm$ 0.029 & 94.1 $\pm$ 7.29 \\ 
Self-(in)consistency & \cellcolor{silver!30}0.188 $\pm$ 0.026 & 0.188 $\pm$ 0.028 & 0.155 $\pm$ 0.05 & 0.15 $\pm$ 0.058 & 1.73 $\pm$ 0.198 & \cellcolor{bronze!30}0.035 $\pm$ 0.028 & 0.081 $\pm$ 0.026 & 89.3 $\pm$ 8.87 \\ 
Binomial NLL & \cellcolor{gold!30}0.176 $\pm$ 0.028 & 0.194 $\pm$ 0.021 & 0.153 $\pm$ 0.044 & 0.149 $\pm$ 0.051 & 1.77 $\pm$ 0.138 & \cellcolor{silver!30}0.034 $\pm$ 0.032 & 0.08 $\pm$ 0.024 & 82.9 $\pm$ 9.5 \\ 
\bottomrule
\end{tabular}
}
    \vspace{1em}
    \caption{ACS Income. Note that this is reported in Table~\ref{tab:fairness_acs} of the main paper body to \textit{2 significant digits} of precision. This was in order to make the font more legible - all results in this section are with 3 significant digits.}
    \label{tab:fairness_acs2}
\end{table}

\begin{table}[h]
    \centering
    \resizebox{\textwidth}{!}{\begin{tabular} {lcccccccc}
\toprule
\textbf{Approach} & \textbf{Error Rate} & \textbf{Statistical Parity} & \textbf{Equalized Odds} & \textbf{Equal Opportunity} & \textbf{Disparate Impact} & \textbf{Predictive Parity} & \textbf{False Positive Rate} & \textbf{Included \%} \\ \midrule
Baseline & 0.265 $\pm$ 0.011 & 0.25 $\pm$ 0.031 & 0.211 $\pm$ 0.04 & 0.168 $\pm$ 0.061 & 0.479 $\pm$ 0.037 & 0.031 $\pm$ 0.03 & 0.194 $\pm$ 0.041 & \cellcolor{gold!30}100.0 $\pm$ 0.0 \\ 
Threshold Optimizer SP & 0.291 $\pm$ 0.016 & \cellcolor{silver!30}0.038 $\pm$ 0.025 & \cellcolor{bronze!30}0.14 $\pm$ 0.033 & 0.129 $\pm$ 0.046 & 0.882 $\pm$ 0.07 & 0.034 $\pm$ 0.038 & \cellcolor{bronze!30}0.077 $\pm$ 0.031 & \cellcolor{gold!30}100.0 $\pm$ 0.0 \\ 
Threshold Optimizer EO & 0.288 $\pm$ 0.017 & 0.168 $\pm$ 0.028 & 0.141 $\pm$ 0.037 & \cellcolor{gold!30}0.069 $\pm$ 0.045 & 0.637 $\pm$ 0.042 & 0.064 $\pm$ 0.039 & 0.139 $\pm$ 0.04 & \cellcolor{gold!30}100.0 $\pm$ 0.0 \\ 
Exponentiated Gradient SP & 0.279 $\pm$ 0.008 & \cellcolor{gold!30}0.035 $\pm$ 0.021 & \cellcolor{gold!30}0.091 $\pm$ 0.052 & \cellcolor{bronze!30}0.088 $\pm$ 0.055 & 0.961 $\pm$ 0.116 & 0.152 $\pm$ 0.039 & \cellcolor{gold!30}0.032 $\pm$ 0.02 & \cellcolor{gold!30}100.0 $\pm$ 0.0 \\ 
Exponentiated Gradient EO & 0.277 $\pm$ 0.014 & 0.176 $\pm$ 0.036 & 0.148 $\pm$ 0.034 & \cellcolor{silver!30}0.073 $\pm$ 0.048 & 0.62 $\pm$ 0.053 & 0.051 $\pm$ 0.04 & 0.147 $\pm$ 0.035 & \cellcolor{gold!30}100.0 $\pm$ 0.0 \\ 
Grid Search SP & 0.291 $\pm$ 0.013 & \cellcolor{bronze!30}0.077 $\pm$ 0.023 & \cellcolor{silver!30}0.124 $\pm$ 0.07 & 0.115 $\pm$ 0.079 & 1.01 $\pm$ 0.255 & 0.139 $\pm$ 0.08 & \cellcolor{silver!30}0.072 $\pm$ 0.042 & \cellcolor{gold!30}100.0 $\pm$ 0.0 \\ 
Grid Search EO & 0.276 $\pm$ 0.014 & 0.221 $\pm$ 0.034 & 0.184 $\pm$ 0.037 & 0.12 $\pm$ 0.054 & 0.557 $\pm$ 0.043 & 0.041 $\pm$ 0.046 & 0.181 $\pm$ 0.041 & \cellcolor{gold!30}100.0 $\pm$ 0.0 \\ 
\midrule
Random & 0.263 $\pm$ 0.012 & 0.239 $\pm$ 0.041 & 0.205 $\pm$ 0.042 & 0.152 $\pm$ 0.077 & 0.495 $\pm$ 0.049 & 0.028 $\pm$ 0.042 & 0.185 $\pm$ 0.043 & 88.4 $\pm$ 8.66 \\ 
Ensemble & \cellcolor{bronze!30}0.248 $\pm$ 0.019 & 0.281 $\pm$ 0.044 & 0.225 $\pm$ 0.054 & 0.21 $\pm$ 0.052 & \cellcolor{silver!30}0.42 $\pm$ 0.036 & \cellcolor{silver!30}0.015 $\pm$ 0.027 & 0.191 $\pm$ 0.042 & 89.6 $\pm$ 7.63 \\ 
Selective Ensemble & \cellcolor{silver!30}0.241 $\pm$ 0.02 & 0.271 $\pm$ 0.033 & 0.236 $\pm$ 0.05 & 0.211 $\pm$ 0.073 & \cellcolor{gold!30}0.412 $\pm$ 0.048 & \cellcolor{bronze!30}0.024 $\pm$ 0.022 & 0.169 $\pm$ 0.044 & 89.0 $\pm$ 8.52 \\ 
Self-(in)consistency & 0.251 $\pm$ 0.02 & 0.249 $\pm$ 0.04 & 0.202 $\pm$ 0.055 & 0.183 $\pm$ 0.064 & 0.447 $\pm$ 0.045 & 0.029 $\pm$ 0.025 & 0.162 $\pm$ 0.051 & 93.3 $\pm$ 7.86 \\ 
Binomial NLL & \cellcolor{gold!30}0.236 $\pm$ 0.024 & 0.257 $\pm$ 0.034 & 0.207 $\pm$ 0.056 & 0.175 $\pm$ 0.071 & \cellcolor{bronze!30}0.436 $\pm$ 0.037 & \cellcolor{gold!30}0.012 $\pm$ 0.021 & 0.166 $\pm$ 0.05 & 87.0 $\pm$ 7.32 \\ 
\bottomrule
\end{tabular}
}
    \vspace{1em}
    \caption{ACS Public Coverage.}
    \label{tab:fairness_acs_public}
\end{table}

\begin{table}[h]
    \centering
    \resizebox{\linewidth}{!}{\begin{tabular} {lcccccccc}
\toprule
\textbf{Approach} & \textbf{Error Rate} & \textbf{Statistical Parity} & \textbf{Equalized Odds} & \textbf{Equal Opportunity} & \textbf{Disparate Impact} & \textbf{Predictive Parity} & \textbf{False Positive Rate} & \textbf{Included \%} \\ \midrule
Baseline & 0.139 $\pm$ 0.002 & 0.168 $\pm$ 0.006 & 0.092 $\pm$ 0.022 & 0.092 $\pm$ 0.022 & 0.291 $\pm$ 0.017 & 0.037 $\pm$ 0.017 & 0.058 $\pm$ 0.005 & \cellcolor{gold!30}100.0 $\pm$ 0.0 \\ 
Threshold Optimizer SP & 0.185 $\pm$ 0.002 & \cellcolor{gold!30}0.007 $\pm$ 0.007 & 0.132 $\pm$ 0.01 & \cellcolor{gold!30}0.024 $\pm$ 0.015 & 0.999 $\pm$ 0.037 & 0.499 $\pm$ 0.013 & 0.132 $\pm$ 0.01 & \cellcolor{gold!30}100.0 $\pm$ 0.0 \\ 
Threshold Optimizer EO & 0.15 $\pm$ 0.002 & 0.122 $\pm$ 0.008 & \cellcolor{gold!30}0.026 $\pm$ 0.018 & \cellcolor{gold!30}0.024 $\pm$ 0.019 & 0.53 $\pm$ 0.028 & 0.271 $\pm$ 0.03 & \cellcolor{gold!30}0.01 $\pm$ 0.006 & \cellcolor{gold!30}100.0 $\pm$ 0.0 \\ 
Exponentiated Gradient SP & 0.15 $\pm$ 0.004 & \cellcolor{silver!30}0.019 $\pm$ 0.009 & 0.288 $\pm$ 0.023 & 0.288 $\pm$ 0.023 & 0.903 $\pm$ 0.045 & 0.335 $\pm$ 0.019 & 0.054 $\pm$ 0.006 & \cellcolor{gold!30}100.0 $\pm$ 0.0 \\ 
Exponentiated Gradient EO & 0.135 $\pm$ 0.002 & 0.134 $\pm$ 0.005 & \cellcolor{silver!30}0.034 $\pm$ 0.008 & \cellcolor{bronze!30}0.028 $\pm$ 0.012 & 0.433 $\pm$ 0.017 & 0.11 $\pm$ 0.032 & \cellcolor{silver!30}0.03 $\pm$ 0.006 & \cellcolor{gold!30}100.0 $\pm$ 0.0 \\ 
Grid Search SP & 0.162 $\pm$ 0.004 & \cellcolor{bronze!30}0.029 $\pm$ 0.009 & 0.359 $\pm$ 0.013 & 0.359 $\pm$ 0.013 & 1.17 $\pm$ 0.055 & 0.412 $\pm$ 0.021 & 0.091 $\pm$ 0.009 & \cellcolor{gold!30}100.0 $\pm$ 0.0 \\ 
Grid Search EO & 0.132 $\pm$ 0.001 & 0.18 $\pm$ 0.005 & \cellcolor{bronze!30}0.075 $\pm$ 0.007 & 0.074 $\pm$ 0.009 & 0.318 $\pm$ 0.008 & \cellcolor{silver!30}0.018 $\pm$ 0.018 & 0.062 $\pm$ 0.005 & \cellcolor{gold!30}100.0 $\pm$ 0.0 \\ 
\midrule
Random & 0.139 $\pm$ 0.002 & 0.168 $\pm$ 0.006 & 0.087 $\pm$ 0.018 & 0.087 $\pm$ 0.018 & 0.293 $\pm$ 0.017 & 0.032 $\pm$ 0.019 & 0.058 $\pm$ 0.005 & 87.1 $\pm$ 6.8 \\ 
Ensemble & \cellcolor{silver!30}0.115 $\pm$ 0.023 & 0.171 $\pm$ 0.01 & 0.09 $\pm$ 0.028 & 0.09 $\pm$ 0.029 & \cellcolor{gold!30}0.24 $\pm$ 0.047 & \cellcolor{bronze!30}0.03 $\pm$ 0.019 & 0.046 $\pm$ 0.009 & 86.6 $\pm$ 11.6 \\ 
Selective Ensemble & \cellcolor{bronze!30}0.129 $\pm$ 0.008 & 0.165 $\pm$ 0.008 & 0.105 $\pm$ 0.026 & 0.105 $\pm$ 0.026 & \cellcolor{bronze!30}0.271 $\pm$ 0.022 & 0.032 $\pm$ 0.021 & 0.049 $\pm$ 0.008 & 96.4 $\pm$ 2.54 \\ 
Self-(in)consistency & 0.134 $\pm$ 0.008 & 0.166 $\pm$ 0.007 & 0.101 $\pm$ 0.029 & 0.101 $\pm$ 0.029 & 0.276 $\pm$ 0.025 & 0.037 $\pm$ 0.021 & 0.052 $\pm$ 0.006 & 97.9 $\pm$ 2.21 \\ 
Binomial NLL & \cellcolor{gold!30}0.106 $\pm$ 0.025 & 0.153 $\pm$ 0.007 & 0.122 $\pm$ 0.053 & 0.122 $\pm$ 0.053 & \cellcolor{silver!30}0.263 $\pm$ 0.042 & \cellcolor{gold!30}0.016 $\pm$ 0.01 & \cellcolor{bronze!30}0.033 $\pm$ 0.015 & 89.6 $\pm$ 7.89 \\ 
\bottomrule
\end{tabular}
}
    \vspace{1em}
    \caption{Adult.}
    \label{tab:fairness_adult}
\end{table}

\begin{table}[h]
    \centering
    \resizebox{\linewidth}{!}{\begin{tabular} {lcccccccc}
\toprule
\textbf{Approach} & \textbf{Error Rate} & \textbf{Statistical Parity} & \textbf{Equalized Odds} & \textbf{Equal Opportunity} & \textbf{Disparate Impact} & \textbf{Predictive Parity} & \textbf{False Positive Rate} & \textbf{Included \%} \\ \midrule
Baseline & 0.092 $\pm$ 0.002 & 0.012 $\pm$ 0.006 & 0.053 $\pm$ 0.027 & 0.053 $\pm$ 0.027 & 0.925 $\pm$ 0.096 & 0.033 $\pm$ 0.016 & 0.004 $\pm$ 0.004 & \cellcolor{gold!30}100.0 $\pm$ 0.0 \\ 
Threshold Optimizer SP & 0.096 $\pm$ 0.002 & \cellcolor{bronze!30}0.006 $\pm$ 0.003 & \cellcolor{gold!30}0.023 $\pm$ 0.014 & \cellcolor{gold!30}0.023 $\pm$ 0.015 & 1.01 $\pm$ 0.062 & 0.038 $\pm$ 0.022 & 0.005 $\pm$ 0.003 & \cellcolor{gold!30}100.0 $\pm$ 0.0 \\ 
Threshold Optimizer EO & 0.097 $\pm$ 0.002 & \cellcolor{bronze!30}0.006 $\pm$ 0.004 & \cellcolor{silver!30}0.025 $\pm$ 0.01 & \cellcolor{silver!30}0.025 $\pm$ 0.011 & 0.975 $\pm$ 0.06 & \cellcolor{gold!30}0.02 $\pm$ 0.018 & 0.003 $\pm$ 0.002 & \cellcolor{gold!30}100.0 $\pm$ 0.0 \\ 
Exponentiated Gradient SP & 0.096 $\pm$ 0.002 & \cellcolor{bronze!30}0.006 $\pm$ 0.004 & 0.032 $\pm$ 0.017 & 0.032 $\pm$ 0.017 & 0.986 $\pm$ 0.067 & 0.035 $\pm$ 0.025 & 0.004 $\pm$ 0.003 & \cellcolor{gold!30}100.0 $\pm$ 0.0 \\ 
Exponentiated Gradient EO & 0.096 $\pm$ 0.002 & \cellcolor{gold!30}0.005 $\pm$ 0.003 & \cellcolor{bronze!30}0.026 $\pm$ 0.015 & \cellcolor{bronze!30}0.026 $\pm$ 0.016 & 0.99 $\pm$ 0.05 & 0.039 $\pm$ 0.025 & 0.005 $\pm$ 0.002 & \cellcolor{gold!30}100.0 $\pm$ 0.0 \\ 
Grid Search SP & 0.097 $\pm$ 0.002 & 0.009 $\pm$ 0.006 & 0.038 $\pm$ 0.021 & 0.038 $\pm$ 0.021 & 0.93 $\pm$ 0.056 & \cellcolor{bronze!30}0.031 $\pm$ 0.031 & 0.004 $\pm$ 0.004 & \cellcolor{gold!30}100.0 $\pm$ 0.0 \\ 
Grid Search EO & 0.097 $\pm$ 0.002 & 0.009 $\pm$ 0.006 & 0.038 $\pm$ 0.021 & 0.038 $\pm$ 0.021 & 0.93 $\pm$ 0.056 & \cellcolor{bronze!30}0.031 $\pm$ 0.031 & 0.004 $\pm$ 0.004 & \cellcolor{gold!30}100.0 $\pm$ 0.0 \\ 
\midrule
Random & 0.092 $\pm$ 0.002 & 0.01 $\pm$ 0.005 & 0.049 $\pm$ 0.028 & 0.048 $\pm$ 0.029 & 0.934 $\pm$ 0.084 & \cellcolor{bronze!30}0.031 $\pm$ 0.017 & 0.003 $\pm$ 0.003 & 91.0 $\pm$ 7.38 \\ 
Ensemble & \cellcolor{gold!30}0.045 $\pm$ 0.026 & \cellcolor{gold!30}0.005 $\pm$ 0.004 & 0.067 $\pm$ 0.05 & 0.067 $\pm$ 0.05 & 0.985 $\pm$ 0.135 & \cellcolor{silver!30}0.021 $\pm$ 0.019 & \cellcolor{gold!30}0.001 $\pm$ 0.0 & 84.3 $\pm$ 8.45 \\ 
Selective Ensemble & 0.077 $\pm$ 0.013 & 0.007 $\pm$ 0.004 & 0.043 $\pm$ 0.035 & 0.042 $\pm$ 0.036 & \cellcolor{bronze!30}0.922 $\pm$ 0.071 & 0.033 $\pm$ 0.021 & \cellcolor{bronze!30}0.002 $\pm$ 0.002 & 95.7 $\pm$ 3.8 \\ 
Self-(in)consistency & \cellcolor{bronze!30}0.076 $\pm$ 0.013 & 0.008 $\pm$ 0.007 & 0.053 $\pm$ 0.044 & 0.053 $\pm$ 0.045 & \cellcolor{gold!30}0.902 $\pm$ 0.109 & 0.038 $\pm$ 0.032 & \cellcolor{bronze!30}0.002 $\pm$ 0.002 & 95.8 $\pm$ 3.82 \\ 
Binomial NLL & \cellcolor{silver!30}0.067 $\pm$ 0.028 & 0.007 $\pm$ 0.005 & 0.043 $\pm$ 0.031 & 0.043 $\pm$ 0.032 & \cellcolor{silver!30}0.908 $\pm$ 0.074 & 0.033 $\pm$ 0.029 & \cellcolor{gold!30}0.001 $\pm$ 0.001 & 92.2 $\pm$ 8.74 \\ 
\bottomrule
\end{tabular}
}
    \vspace{1em}
    \caption{Bank.}
    \label{tab:fairness_bank}
\end{table}

\begin{table}[h]
    \centering
    \resizebox{\linewidth}{!}{\begin{tabular} {lcccccccc}
\toprule
\textbf{Approach} & \textbf{Error Rate} & \textbf{Statistical Parity} & \textbf{Equalized Odds} & \textbf{Equal Opportunity} & \textbf{Disparate Impact} & \textbf{Predictive Parity} & \textbf{False Positive Rate} & \textbf{Included \%} \\ \midrule
Baseline & 0.316 $\pm$ 0.012 & 0.194 $\pm$ 0.027 & 0.188 $\pm$ 0.05 & 0.183 $\pm$ 0.054 & 0.542 $\pm$ 0.055 & 0.075 $\pm$ 0.051 & 0.125 $\pm$ 0.039 & \cellcolor{gold!30}100.0 $\pm$ 0.0 \\ 
Threshold Optimizer SP & 0.344 $\pm$ 0.015 & \cellcolor{silver!30}0.042 $\pm$ 0.025 & 0.112 $\pm$ 0.047 & \cellcolor{bronze!30}0.074 $\pm$ 0.052 & 0.963 $\pm$ 0.107 & 0.227 $\pm$ 0.067 & 0.09 $\pm$ 0.053 & \cellcolor{gold!30}100.0 $\pm$ 0.0 \\ 
Threshold Optimizer EO & 0.35 $\pm$ 0.016 & 0.065 $\pm$ 0.031 & \cellcolor{silver!30}0.074 $\pm$ 0.048 & \cellcolor{silver!30}0.066 $\pm$ 0.052 & 0.82 $\pm$ 0.084 & 0.163 $\pm$ 0.067 & \cellcolor{bronze!30}0.04 $\pm$ 0.028 & \cellcolor{gold!30}100.0 $\pm$ 0.0 \\ 
Exponentiated Gradient SP & 0.329 $\pm$ 0.013 & \cellcolor{gold!30}0.041 $\pm$ 0.025 & \cellcolor{gold!30}0.051 $\pm$ 0.027 & \cellcolor{gold!30}0.038 $\pm$ 0.033 & 0.902 $\pm$ 0.069 & 0.161 $\pm$ 0.05 & \cellcolor{gold!30}0.028 $\pm$ 0.018 & \cellcolor{gold!30}100.0 $\pm$ 0.0 \\ 
Exponentiated Gradient EO & 0.332 $\pm$ 0.014 & 0.083 $\pm$ 0.028 & \cellcolor{bronze!30}0.085 $\pm$ 0.046 & 0.084 $\pm$ 0.047 & 0.792 $\pm$ 0.071 & 0.164 $\pm$ 0.066 & \cellcolor{silver!30}0.038 $\pm$ 0.025 & \cellcolor{gold!30}100.0 $\pm$ 0.0 \\ 
Grid Search SP & 0.327 $\pm$ 0.015 & \cellcolor{bronze!30}0.047 $\pm$ 0.058 & 0.112 $\pm$ 0.06 & 0.097 $\pm$ 0.063 & 0.991 $\pm$ 0.182 & 0.176 $\pm$ 0.043 & 0.07 $\pm$ 0.042 & \cellcolor{gold!30}100.0 $\pm$ 0.0 \\ 
Grid Search EO & 0.323 $\pm$ 0.012 & 0.213 $\pm$ 0.03 & 0.224 $\pm$ 0.059 & 0.223 $\pm$ 0.06 & 0.517 $\pm$ 0.054 & 0.088 $\pm$ 0.063 & 0.135 $\pm$ 0.034 & \cellcolor{gold!30}100.0 $\pm$ 0.0 \\ 
\midrule
Random & 0.317 $\pm$ 0.014 & 0.181 $\pm$ 0.025 & 0.168 $\pm$ 0.04 & 0.154 $\pm$ 0.052 & 0.573 $\pm$ 0.051 & 0.063 $\pm$ 0.067 & 0.12 $\pm$ 0.043 & 86.3 $\pm$ 6.72 \\ 
Ensemble & 0.305 $\pm$ 0.011 & 0.201 $\pm$ 0.033 & 0.206 $\pm$ 0.06 & 0.197 $\pm$ 0.076 & \cellcolor{bronze!30}0.5 $\pm$ 0.09 & \cellcolor{bronze!30}0.056 $\pm$ 0.048 & 0.127 $\pm$ 0.022 & 87.8 $\pm$ 8.68 \\ 
Selective Ensemble & \cellcolor{silver!30}0.288 $\pm$ 0.015 & 0.211 $\pm$ 0.047 & 0.213 $\pm$ 0.07 & 0.202 $\pm$ 0.088 & \cellcolor{silver!30}0.498 $\pm$ 0.107 & \cellcolor{silver!30}0.054 $\pm$ 0.046 & 0.126 $\pm$ 0.032 & 88.0 $\pm$ 7.52 \\ 
Self-(in)consistency & \cellcolor{gold!30}0.287 $\pm$ 0.019 & 0.213 $\pm$ 0.033 & 0.219 $\pm$ 0.071 & 0.216 $\pm$ 0.074 & \cellcolor{gold!30}0.485 $\pm$ 0.067 & 0.07 $\pm$ 0.06 & 0.121 $\pm$ 0.034 & 86.9 $\pm$ 9.61 \\ 
Binomial NLL & \cellcolor{bronze!30}0.292 $\pm$ 0.018 & 0.206 $\pm$ 0.033 & 0.199 $\pm$ 0.058 & 0.196 $\pm$ 0.061 & \cellcolor{bronze!30}0.5 $\pm$ 0.074 & \cellcolor{gold!30}0.052 $\pm$ 0.047 & 0.123 $\pm$ 0.035 & 88.7 $\pm$ 8.4 \\ 
\bottomrule
\end{tabular}
}
    \vspace{1em}
    \caption{COMPAS.}
    \label{tab:fairness_compas}
\end{table}

\begin{table}[h]
    \centering
    \resizebox{\linewidth}{!}{\begin{tabular} {lcccccccc}
\toprule
\textbf{Approach} & \textbf{Error Rate} & \textbf{Statistical Parity} & \textbf{Equalized Odds} & \textbf{Equal Opportunity} & \textbf{Disparate Impact} & \textbf{Predictive Parity} & \textbf{False Positive Rate} & \textbf{Included \%} \\ \midrule
Baseline & 0.251 $\pm$ 0.025 & 0.129 $\pm$ 0.06 & 0.147 $\pm$ 0.064 & \cellcolor{bronze!30}0.129 $\pm$ 0.07 & 1.72 $\pm$ 0.35 & 0.082 $\pm$ 0.048 & 0.094 $\pm$ 0.067 & \cellcolor{gold!30}100.0 $\pm$ 0.0 \\ 
Threshold Optimizer SP & 0.258 $\pm$ 0.026 & \cellcolor{silver!30}0.081 $\pm$ 0.059 & 0.175 $\pm$ 0.082 & 0.152 $\pm$ 0.097 & \cellcolor{gold!30}1.08 $\pm$ 0.469 & 0.187 $\pm$ 0.11 & 0.078 $\pm$ 0.062 & \cellcolor{gold!30}100.0 $\pm$ 0.0 \\ 
Threshold Optimizer EO & 0.255 $\pm$ 0.029 & 0.155 $\pm$ 0.081 & 0.166 $\pm$ 0.077 & 0.151 $\pm$ 0.065 & 1.75 $\pm$ 0.423 & 0.105 $\pm$ 0.105 & 0.124 $\pm$ 0.092 & \cellcolor{gold!30}100.0 $\pm$ 0.0 \\ 
Exponentiated Gradient SP & 0.264 $\pm$ 0.025 & \cellcolor{gold!30}0.065 $\pm$ 0.056 & \cellcolor{gold!30}0.11 $\pm$ 0.066 & \cellcolor{gold!30}0.089 $\pm$ 0.063 & \cellcolor{silver!30}1.29 $\pm$ 0.371 & 0.112 $\pm$ 0.112 & \cellcolor{silver!30}0.063 $\pm$ 0.068 & \cellcolor{gold!30}100.0 $\pm$ 0.0 \\ 
Exponentiated Gradient EO & 0.255 $\pm$ 0.029 & 0.155 $\pm$ 0.081 & 0.166 $\pm$ 0.077 & 0.151 $\pm$ 0.065 & 1.75 $\pm$ 0.423 & 0.105 $\pm$ 0.105 & 0.124 $\pm$ 0.092 & \cellcolor{gold!30}100.0 $\pm$ 0.0 \\ 
Grid Search SP & 0.269 $\pm$ 0.023 & 0.136 $\pm$ 0.079 & 0.175 $\pm$ 0.073 & 0.16 $\pm$ 0.072 & \cellcolor{bronze!30}1.46 $\pm$ 0.602 & 0.086 $\pm$ 0.068 & 0.111 $\pm$ 0.078 & \cellcolor{gold!30}100.0 $\pm$ 0.0 \\ 
Grid Search EO & 0.259 $\pm$ 0.038 & 0.15 $\pm$ 0.094 & 0.197 $\pm$ 0.108 & 0.186 $\pm$ 0.109 & 1.67 $\pm$ 0.443 & 0.109 $\pm$ 0.1 & 0.108 $\pm$ 0.073 & \cellcolor{gold!30}100.0 $\pm$ 0.0 \\ 
\midrule
Random & 0.25 $\pm$ 0.026 & 0.116 $\pm$ 0.057 & \cellcolor{silver!30}0.139 $\pm$ 0.075 & \cellcolor{silver!30}0.107 $\pm$ 0.091 & 1.64 $\pm$ 0.315 & \cellcolor{silver!30}0.068 $\pm$ 0.04 & 0.087 $\pm$ 0.06 & 92.5 $\pm$ 8.25 \\ 
Ensemble & \cellcolor{bronze!30}0.222 $\pm$ 0.046 & 0.125 $\pm$ 0.076 & 0.155 $\pm$ 0.088 & 0.142 $\pm$ 0.1 & 2.04 $\pm$ 0.753 & 0.069 $\pm$ 0.077 & 0.077 $\pm$ 0.054 & 85.4 $\pm$ 9.95 \\ 
Selective Ensemble & \cellcolor{gold!30}0.214 $\pm$ 0.045 & 0.122 $\pm$ 0.064 & 0.176 $\pm$ 0.097 & 0.157 $\pm$ 0.113 & 1.93 $\pm$ 0.496 & 0.082 $\pm$ 0.084 & 0.073 $\pm$ 0.047 & 88.4 $\pm$ 8.57 \\ 
Self-(in)consistency & \cellcolor{silver!30}0.216 $\pm$ 0.046 & 0.109 $\pm$ 0.063 & 0.148 $\pm$ 0.084 & 0.142 $\pm$ 0.092 & 1.85 $\pm$ 0.481 & \cellcolor{silver!30}0.068 $\pm$ 0.087 & \cellcolor{gold!30}0.062 $\pm$ 0.042 & 90.3 $\pm$ 8.45 \\ 
Binomial NLL & 0.232 $\pm$ 0.023 & \cellcolor{bronze!30}0.107 $\pm$ 0.049 & \cellcolor{bronze!30}0.143 $\pm$ 0.085 & 0.132 $\pm$ 0.091 & 1.68 $\pm$ 0.351 & \cellcolor{gold!30}0.057 $\pm$ 0.034 & \cellcolor{bronze!30}0.065 $\pm$ 0.054 & 92.6 $\pm$ 7.86 \\ 
\bottomrule
\end{tabular}
}
    \vspace{1em}
    \caption{German.}
    \label{tab:fairness_german}
\end{table}

\clearpage

\section{Binary Calibration with Different Sized Groups}\label{app:different_sizes}

We find that bigger groups (a smaller number of groups) exhibit more identifiable patterns with a positive correlation whereas smaller groups (a bigger number of groups) exhibit slightly more noisy behavior.

\begin{figure}[h!]
    \centering
    \includegraphics[width=\linewidth]{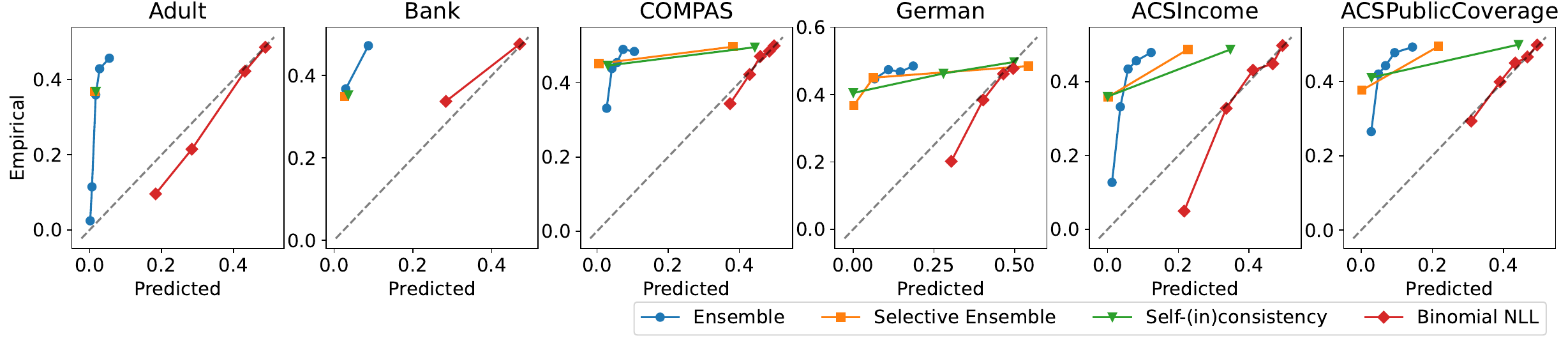}
    \caption{For five groups assembled by predicted uncertainty, we plot the average predicted uncertainty against the empirical standard deviation of the outcomes.}
    \label{fig:binary_calibration5}
\end{figure}

\begin{figure}[h!]
    \centering
    \includegraphics[width=\linewidth]{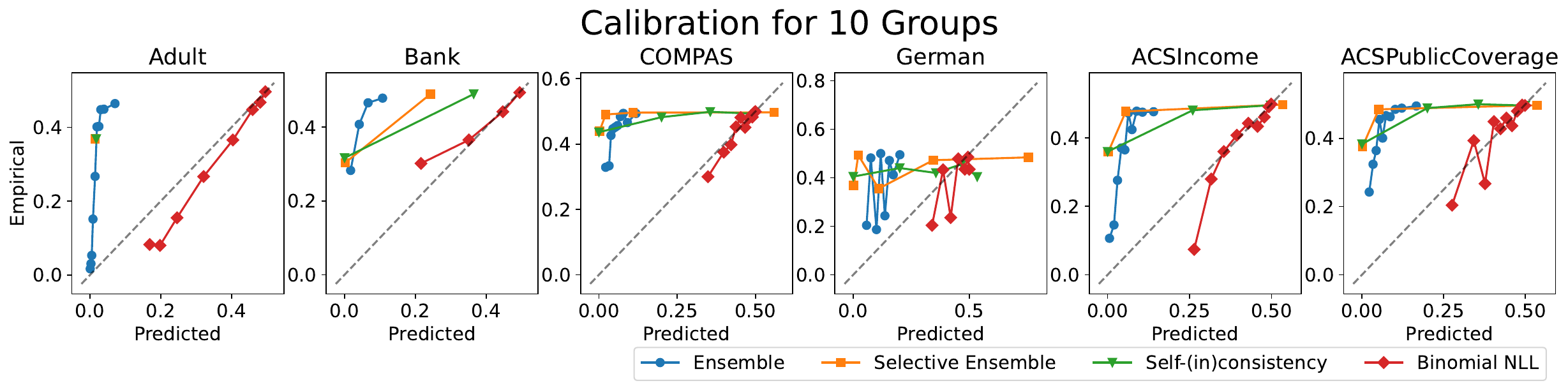}
    \caption{For ten groups assembled by predicted uncertainty, we plot the average predicted uncertainty against the empirical standard deviation of the outcomes.}
    \label{fig:binary_calibration10}
\end{figure}

\begin{figure}[h!]
    \centering
    \includegraphics[width=\linewidth]{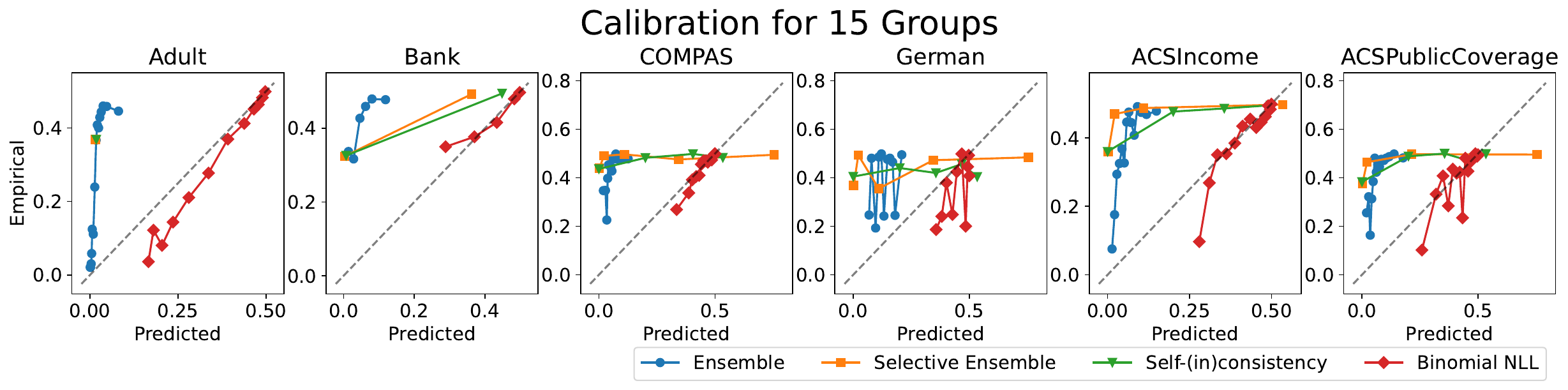}
    \caption{For fifteen groups assembled by predicted uncertainty, we plot the average predicted uncertainty against the empirical standard deviation of the outcomes.}
    \label{fig:binary_calibration15}
\end{figure}

\begin{figure}[h!]
    \centering
    \includegraphics[width=\linewidth]{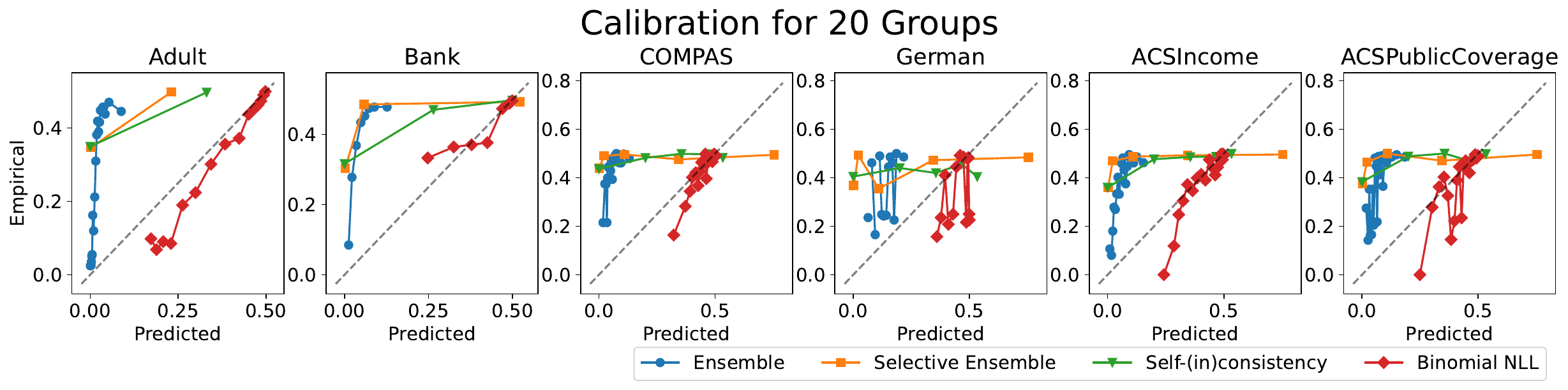}
    \caption{For twenty groups assembled by predicted uncertainty, we plot the average predicted uncertainty against the empirical standard deviation of the outcomes.}
    \label{fig:binary_calibration20}
\end{figure}

\begin{figure}[h!]
    \centering
    \includegraphics[width=\linewidth]{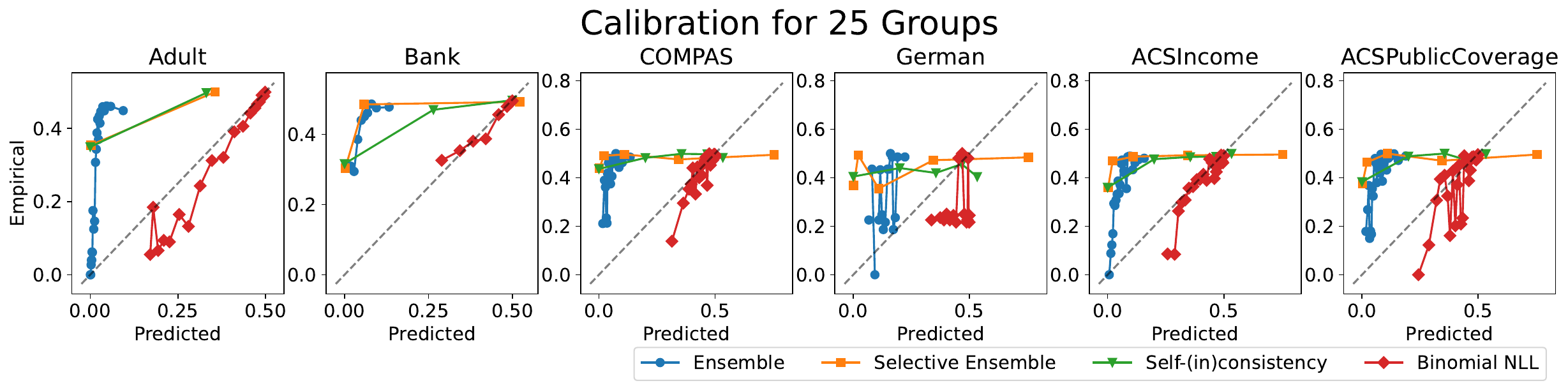}
    \caption{For twenty five groups assembled by predicted uncertainty, we plot the average predicted uncertainty against the empirical standard deviation of the outcomes.}
    \label{fig:binary_calibration25}
\end{figure}

\begin{figure}[h!]
    \centering
    \includegraphics[width=\linewidth]{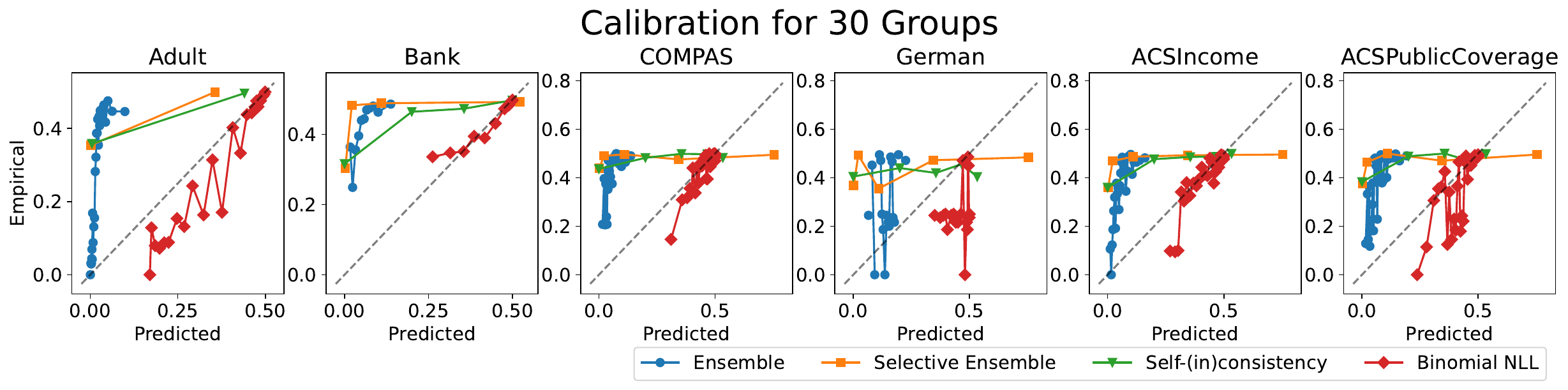}
    \caption{For thirty groups assembled by predicted uncertainty, we plot the average predicted uncertainty against the empirical standard deviation of the outcomes.}
    \label{fig:binary_calibration30}
\end{figure}

\clearpage
\section{Notes on XGBoost and Customizing the Loss}\label{app:xgboost}
XGBoost optimizes a loss using its gradient $g_i$ for first-order information and its Hessian $h_i$ for second-order information in a relatively \href{https://en.wikipedia.org/wiki/Hessian_matrix#Use_in_optimization}{standard}, higher order optimization framework (see Section 2.2. in \cite{chen2016xgboost}). Intuitively, $g_i$ indicates the direction for improving model predictions, and $h_i$ is useful for determining the curvature or ``rate of change'' of the loss. During XGBoost tree construction, each tree is built iteratively in order to correct the ``residuals'' left by previous trees. Here, ``residuals'' refer to the differences between the observed values of the target variable and the values predicted by the model e.g. the errors in the predictions made by the model. We let
\begin{align}
w^*_j = -\frac{\sum_{i \in I_j} g_i}{\sum_{i \in I_j} h_i + \lambda},
\end{align}
where $ w^*_j $ is the optimal weight for the $ j $-th leaf, $ I_j $ represents the set of instances in leaf $ j $, and $ \lambda $ is a regularizer. Formulating it in this way is the classic balancing act - we want to reduce the loss, but we regularize against the complexity of the model. Then, when we want to evaluate potential splits during tree construction, the ``gain'' from a split is calculated as:
\begin{align}
\text{Gain} = \frac{1}{2} \left[\frac{(\sum_{i \in I_L} g_i)^2}{\sum_{i \in I_L} h_i + \lambda} + \frac{(\sum_{i \in I_R} g_i)^2}{\sum_{i \in I_R} h_i + \lambda} - \frac{(\sum_{i \in I} g_i)^2}{\sum_{i \in I} h_i + \lambda}\right] - \gamma,
\end{align}
where $ I_L $ and $ I_R $ are the instance sets of the left and right child nodes post-split. Thus, specifying a custom loss for XGBoost simply requires deriving the first order gradient and second order hessian for a given loss function with arbitrary outputs. 

\clearpage
\section{Fairness Metrics}\label{app:fairness_metrics}
Below we provide formulas for the fairness metrics included in our benchmark. Let $X$ denote the set of features, $Y$ the true label, $\hat{Y}$ the predicted label, and $A$ the protected attribute (e.g., race, gender). Note that we generally report these as \textit{metrics}, which is to say we quantify the degree of fairness by calculating the absolute difference between the two sides of the equality in each definition. For disparate impact, we report $\delta$ directly.

\begin{definition}[Statistical Parity]
\begin{align}
P(\hat{Y} = 1 \mid A = a) = P(\hat{Y} = 1 \mid A = a') \quad \forall a, a'
\end{align}
i.e. the probability of receiving a positive outcome should be the same across all $A$.
\end{definition}

\begin{definition}[Equalized Odds]
\begin{align}
P(\hat{Y} = 1 \mid A = a, Y = y) = P(\hat{Y} = 1 \mid A = a', Y = y) \quad \forall a, a', y
\end{align}
i.e. the prediction outcome is conditionally independent of the protected attribute $A$ given $Y$.
\end{definition}

\begin{definition}[Equal Opportunity]
\begin{align}
P(\hat{Y} = 1 \mid A = a, Y = 1) = P(\hat{Y} = 1 \mid A = a', Y = 1) \quad \forall a, a'
\end{align}
i.e. the true positive rate is the same across all $A$.
\end{definition}

\begin{definition}[Disparate Impact]
\begin{align}
\frac{P(\hat{Y} = 1 \mid A = a)}{P(\hat{Y} = 1 \mid A = a')} \geq \delta \quad \text{for some threshold } \delta
\end{align}
i.e. the ratio of positive outcomes between groups remains close to 1 (typically $\delta = 0.8$).
\end{definition}

\begin{definition}[Predictive Parity]
\begin{align}
P(Y = 1 \mid \hat{Y} = 1, A = a) = P(Y = 1 \mid \hat{Y} = 1, A = a') \quad \forall a, a'
\end{align}
i.e. precision (the probability of a true positive given a positive prediction) is the same across all $A$.
\end{definition}

\begin{definition}[False Positive Rate Equality]
\begin{align}
P(\hat{Y} = 1 \mid A = a, Y = 0) = P(\hat{Y} = 1 \mid A = a', Y = 0) \quad \forall a, a'
\end{align}
i.e. the false positive rate is the same across all $A$.
\end{definition}

\end{document}